\documentclass[11pt]{article}

\usepackage[preprint]{acl}

\usepackage{times}
\usepackage{latexsym}

\usepackage[T1]{fontenc}

\usepackage[utf8]{inputenc}

\usepackage{microtype}

\usepackage{inconsolata}

\usepackage{graphicx}

\usepackage{url}
\usepackage{amsmath}
\usepackage{amsfonts}
\usepackage{amssymb}
\usepackage{hyperref}
\usepackage{booktabs}
\usepackage{multirow}
\usepackage{twemojis}
\usepackage{tcolorbox}
\tcbuselibrary{skins}
\usepackage[normalem]{ulem}
\usepackage{colortbl}

\definecolor{revcolor}{HTML}{C1440E}

\definecolor{revcolorb}{HTML}{0072B2}

\usepackage[subtle]{savetrees}

\usepackage{tabularray}
\UseTblrLibrary{booktabs}

\usepackage{silence}
\definecolor{yang}{HTML}{2da02c}

\newcommand{\textmod}{\scalebox{1.25}{\twemoji{pencil}}}
\newcommand{\audiomod}{\scalebox{1.25}{\twemoji{studio microphone}}}

\newcommand{\videomod}{\scalebox{1.25}{\twemoji{film frames}}}

\newcommand{\email}{\scalebox{1}{\twemoji{email}}}

\definecolor{rowgray}{gray}{0.97}
\definecolor{lightblue}{RGB}{204,229,255}
\definecolor{darkblue}{HTML}{0d47a1}
\definecolor{lightred}{RGB}{255,204,204}
\definecolor{TodoColor}{rgb}{1,0.7,0.6}
\definecolor{TodoColor2}{rgb}{0.7,0.7,0.9}
\definecolor{TodoColor3}{rgb}{0.5,0.8,0.5}
\definecolor{coquelicot}{rgb}{1.0, 0.22, 0.0}

\definecolor{topone}{HTML}{89E0CD}
\definecolor{toptwo}{HTML}{A9EADF}
\definecolor{topthree}{HTML}{C8F3EA}
\definecolor{bottomone}{HTML}{C7AEBB}
\definecolor{bottomtwo}{HTML}{D8C4CE}
\definecolor{bottomthree}{HTML}{E8DDE3}
\newcommand{\bestA}[1]{\cellcolor{topone}\textbf{#1}}
\newcommand{\bestB}[1]{\cellcolor{toptwo}#1}
\newcommand{\bestC}[1]{\cellcolor{topthree}#1}
\newcommand{\worstA}[1]{\cellcolor{bottomone}#1}
\newcommand{\worstB}[1]{\cellcolor{bottomtwo}#1}
\newcommand{\worstC}[1]{\cellcolor{bottomthree}#1}

\definecolor{purpleGreek}{HTML}{800080}
\definecolor{blueGreek}{HTML}{4891e7}
\definecolor{redGreek}{HTML}{d9381e}
\definecolor{greenGreek}{HTML}{088f8f}
\definecolor{yellowGreek}{HTML}{E0BC00}
\definecolor{orangeGreek}{HTML}{FFA500}

\newcommand{\uom}{{\color{purpleGreek}\boldsymbol{\mu}}}
\newcommand{\uos}{{\color{blueGreek}\boldsymbol{\sigma}}}
\newcommand{\du}{{\color{redGreek}\boldsymbol{\delta}}}
\newcommand{\uoe}{{\color{greenGreek}\boldsymbol{\epsilon}}}
\newcommand{\mc}{{\color{yellowGreek}\boldsymbol{\kappa}}}
\newcommand{\sufe}{{\color{orangeGreek}\boldsymbol{\zeta}}}

\title{Reading Between the Frames of Multimodal Drivelology}

\author{
  \parbox{\textwidth}{\centering
    \textbf{Yang Wang}$\hspace{.1em}^{\uom}$,
    \textbf{Yanan Ma}$\hspace{.1em}^{\mc}$,
    \textbf{Zi Yan Chang}$\hspace{.1em}^{\uos}$,
    \textbf{Tyler Loakman}$\hspace{.1em}^{\uos}$,
    \textbf{Chi-Li Chen}$\hspace{.1em}^{\uos}$,
    \textbf{Chia-Yi Hsiao}$\hspace{.1em}^{\du}$,
  \\ \vspace{0.1cm}
    \textbf{Yeqi Jin}$\hspace{.1em}^{\uom}$,
    \textbf{Yiqi Liu}$\hspace{.1em}^{\uom}$,
    \textbf{Aline Villavicencio}$\hspace{.1em}^{\uoe}$$\hspace{.1em}^{\uos}$,
    \textbf{Chenghao Xiao}$\hspace{.1em}^{\sufe}$,
    \textbf{Chenghua Lin}$\hspace{.1em}^{\uom}$$\hspace{.1em}^{\email}$
  \\ \vspace{0.4cm}
    \textnormal{$\hspace{.1em}^{\uom}$University of Manchester},
    \textnormal{$\hspace{.1em}^{\sufe}$Shanghai University of Finance and Economics},\\
    \textnormal{$\hspace{.1em}^{\du}$Durham University},
    \textnormal{$\hspace{.1em}^{\uos}$University of Sheffield},
    \textnormal{$\hspace{.1em}^{\uoe}$University of Exeter},
    \textnormal{$\hspace{.1em}^{\mc}$Mayo Clinic}
  \\ \vspace{0.4cm}
    \raisebox{-0.4ex}{\includegraphics[height=1em]{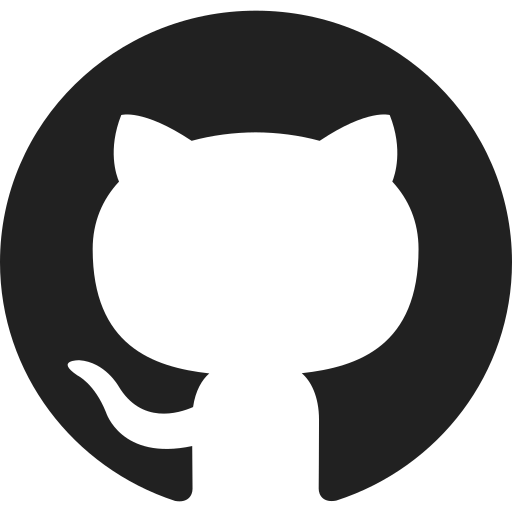}}\hspace{0.3em}\href{https://github.com/extraordinarylab/drivel-hub-plus}{Codebase}\hspace{0.8em}
    \raisebox{-0.4ex}{\includegraphics[height=1em]{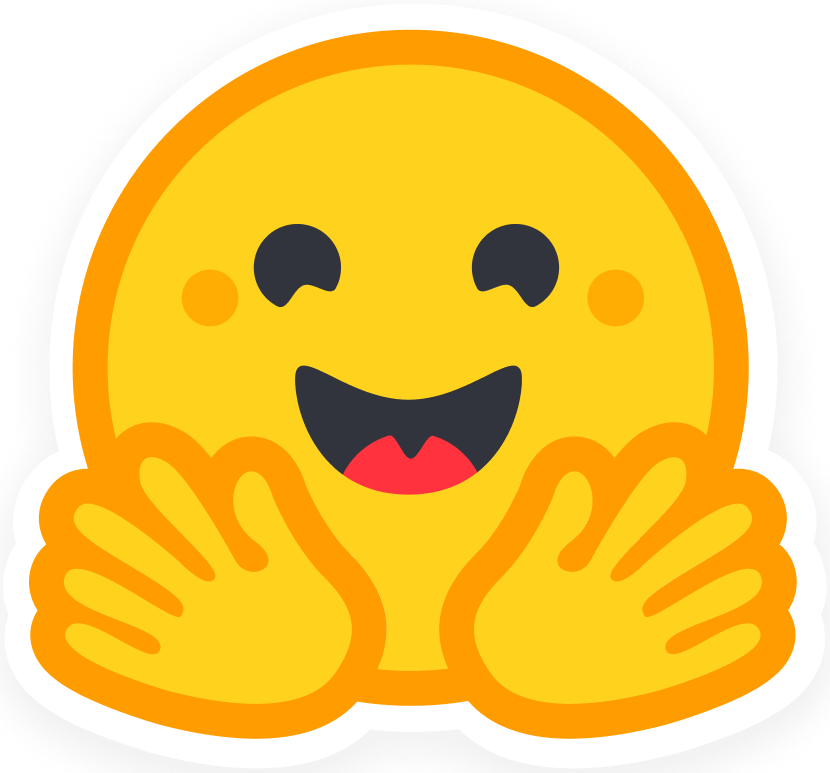}}\hspace{0.3em}\href{https://huggingface.co/datasets/extraordinarylab/drivel-hub-plus}{Dataset}\hspace{0.8em}
    \raisebox{-0.4ex}{\includegraphics[height=1em]{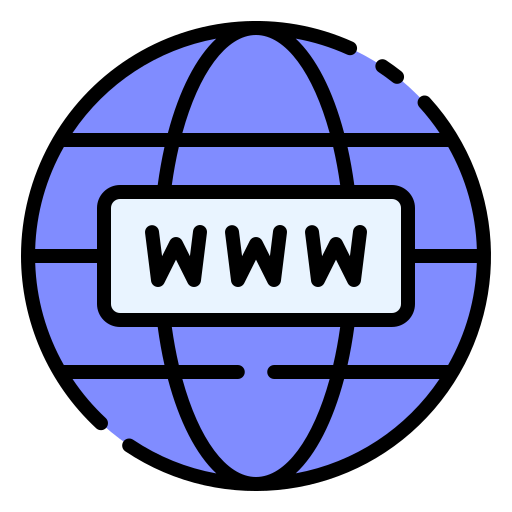}}\hspace{0.3em}\href{https://extraordinarylab.github.io/drivel-hub-plus}{Website}\hspace{0.8em}
  }
}

\begin{document}
\maketitle

\newcommand{\corresponding}[1]{
  \begingroup
  \renewcommand{\thefootnote}{}
  \makeatletter
  \def\@makefnmark{}
  \makeatother
  \begin{NoHyper}
    \footnotetext{#1}%
  \end{NoHyper}
  \endgroup
}
\corresponding{\email\hspace{.1em} The corresponding author is Chenghua Lin, who can be contacted at \url{chenghua.lin@manchester.ac.uk}.}

\newcommand{\aclwarning}[1]{
  \begingroup
  \renewcommand{\thefootnote}{}
  \makeatletter
  \def\@makefnmark{}
  \makeatother
  \begin{NoHyper}
    \footnotetext{#1}%
  \end{NoHyper}
  \endgroup
}
\aclwarning{\textcolor{purple}{\textbf{Content Warning:} This paper analyses social media videos and annotations with potentially offensive or sensitive content. The examples are included for academic analysis and do not reflect the views of the authors.}}

\begin{abstract}
Social media videos often communicate meanings that go beyond their visible actions, captions, or speech. A mundane clip may become humorous, ironic, or satire only through the interaction of multimodal cues and cultural context, making such content a difficult test case for video-language models. In this paper, we introduce \textit{DrivelHub+}, a benchmark for evaluating whether models can infer the implicit, non-linear, and rhetorically layered meanings of social media videos that appear nonsensical on the surface but convey deliberate pragmatic meanings. DrivelHub+ consists of 1,000 videos collected from social media, each annotated with a human-written implicit narrative explanation. Unlike conventional video understanding tasks focused on recognition or description, we present a benchmark that targets contextual multimodal reasoning. We evaluate current video-language models from two perspectives: explanation, where models must explain the pragmatic comprehension of a video in natural language; and representation, where we adapt reasoning-as-retrieval to test whether model representations align videos with their corresponding implicit narratives in both video-to-text and text-to-video retrieval. Our benchmark provides a diagnostic setting for measuring the gap between multimodal perception and pragmatic comprehension, asking whether current models can move beyond describing what is shown to inferring what is meant.
\end{abstract}



\section{Introduction}

\begin{figure*}[t]
    \centering
    \includegraphics[width=0.9\linewidth]{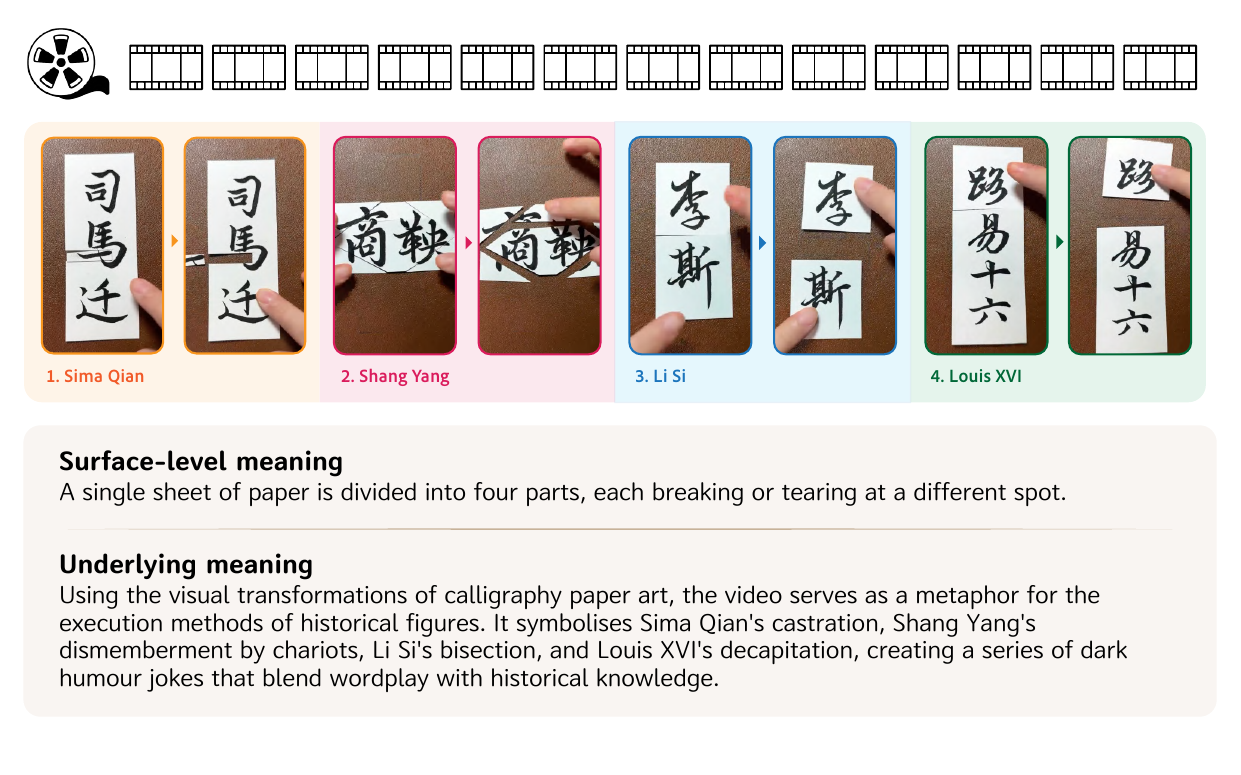}
    \caption{
    Example of a drivelological video where surface-level paper transformations encode historically associated punishments through wordplay, visual transformation, and cultural knowledge.
    }
    \label{fig:example}
\end{figure*}

Short-form social media videos rarely convey surface-level meaning alone. A TikTok, Instagram Reel, or YouTube Short may appear mundane at first glance: a person performs an everyday action, says an ordinary sentence, or adds a short caption, yet its communicative force may depend on a mismatch between what is shown and what is meant \cite{li2026vimubenchmarkingvideometaphorical}. Humour, irony, or satire can emerge from a delayed punchline, a visual contradiction, a deadpan expression, an ironic caption, or an implicit cultural reference. In such cases, faithful surface description is insufficient; understanding requires inferring the implicit narrative produced through the interaction of modalities \cite{xie2024funqa, long2025adsqa, li2026vimubenchmarkingvideometaphorical}.

\textit{Drivelology} \citep{wang2025drivel} is a form of \textit{nonsense with depth}. Drivelological expressions are syntactically coherent but pragmatically paradoxical, rhetorically subversive, or emotionally loaded. They may appear trivial or absurd on the surface while encoding a hidden humour, satire, or social observation. The original, text-based DrivelHub benchmark demonstrated that even strong LLMs often confuse such layered rhetorical structures with shallow nonsense \cite{wang2025drivel}. However, much drivelological communication online is multimodal rather than purely textual: the implicit meaning may be carried by the contrast between speech and action, the timing of a cut, the tone of delivery, the relation between caption and scene, or culturally situated visual knowledge.

This makes drivelological video understanding a difficult test case for current Multimodal Large Language Models (MLLMs) and Video Large Language Models (Video LLMs). The same action, utterance, or caption supports different readings depending on what it is paired with in another modality, so the difficulty is not object recognition or temporal grounding but pragmatic interpretation: a model may recognise the people, objects, actions, captions, and speech while still missing why these elements jointly become meaningful.

Most existing video benchmarks do not test this capability. They mainly focus on perceptual and descriptive tasks: action recognition \cite{zhou2024mlvu}, temporal localisation \cite{lei2020tvqa+}, object tracking \cite{wang2024omnivid, fu2025mme}, question answering \cite{mangalam2023egoschema, Maaz2024VideoChatGPT, Nagrani_2025_ICCV, wang2025lvbench, dai2025towards}, and captioning \cite{fu2025video, fu2026video}. These test whether a model can identify what explicitly occurs. Drivelological videos ask a complementary question: \textit{what does the video imply, and how is that implication constructed from multimodal evidence?}

Recent work has begun to evaluate non-literal and socially grounded multimodal understanding, including sarcasm, irony, intent, metaphor, humour, and social reasoning \cite{farabi2024survey, kong2025siv, kang2025sarcasm, goel2025target, zhang2026cfms, li2026vimubenchmarkingvideometaphorical}. Some of this work is image-based, while a growing subset focuses on video. Video-oriented benchmarks such as MUStReason \cite{saha2025mustreason}, SIV-Bench \cite{kong2025siv}, and ViMU \cite{li2026vimubenchmarkingvideometaphorical} represent an important shift toward pragmatic multimodal interpretation. However, many are framed as classification, multiple-choice selection, or category-level recognition. Such formats support controlled evaluation but understate the interpretive challenge, where understanding lies less in assigning a label than in explaining the mechanism by which humour, satire, or stance is produced.

In this paper, we introduce the \textit{DrivelHub+} dataset and use it to construct a benchmark of drivelological videos collected from social media. Each video is paired with a human-written annotation describing the implicit meaning behind its seemingly nonsensical surface, and the task is to explain how an apparently trivial or absurd video becomes meaningful through visual, textual, auditory, temporal, and cultural cues. Figure~\ref{fig:example} illustrates this challenge: the visible action can be described simply as paper being divided or torn, but the implicit meaning depends on mapping these visual transformations to historical punishments through cultural knowledge and metaphorical reasoning. Additional examples illustrating modality-specific and cross-modal reasoning challenges appear in Figures~\ref{fig:example_text}-\ref{fig:example_text_audio_vision}.

We evaluate this capability from two complementary perspectives. First, we prompt Video LLMs to explain the implicit meaning of each drivelological video in natural language, testing whether they can move beyond surface description and articulate the interpretation, including the relevant joke, critique, affective implication, or pragmatic mechanism. Second, we adapt reasoning-as-retrieval \cite{xiao2024rar} to test whether model representations support alignment between videos and their implicit narratives. We formulate this as bidirectional cross-modal retrieval: \textit{video-to-text} retrieval asks a model to retrieve the correct implicit narrative for a given video, while \textit{text-to-video} retrieval asks it to retrieve the corresponding video for a given narrative. Explanation tests whether a model can articulate the interpretation; retrieval tests whether the video and its interpretation are organised together in representation space rather than matched only by surface actions, objects, captions, or lexical overlap.

Our contributions are:
\begin{itemize}
    \item We formulate \textit{multimodal drivelological explanation} as a task for evaluating implicit pragmatic meaning in short-form social-media videos.
    \item We introduce \textit{DrivelHub+}, a dataset and benchmark of 1,000 videos with human-written implicit narrative annotations.
    \item We propose a two-level evaluation protocol including explanation and bidirectional video-text retrieval, probing both articulated interpretation and representation-level alignment.
    \item We show that current models often recognise surface content but struggle to explain or represent the deeper pragmatic meanings produced by multimodal incongruity, wordplay, cultural knowledge, and social context.
\end{itemize}

\section{Related Work}

Video-language benchmarks cover video captioning \cite{fu2025video, fu2026video}, video question answering \cite{mangalam2023egoschema, Maaz2024VideoChatGPT, wang2025lvbench}, temporal localisation \cite{lei2020tvqa+}, action-centric reasoning \cite{zhou2024mlvu}, event understanding \cite{liang2025videvent}, and long-context video reasoning \cite{wu2024longvideobench}. These measure whether models recognise actions, track events, answer questions about visible content, or reason over extended temporal contexts.
Moving beyond surface perception, benchmarks such as SIV-Bench \cite{kong2025siv}, WildVideo \cite{yang2025wildvideo}, VideoMind \cite{yang2025videomind}, ImplicitQA \cite{swetha2025implicitqa}, MAVERIX \cite{xie2026maverix}, GODBench \cite{lei2025godbench}, and ViMU \cite{li2026vimubenchmarkingvideometaphorical} evaluate social interaction, theory-of-mind, intent grounding, implicit causality, audio-visual alignment, metaphorical understanding, and non-literal cultural commentary. 
However, they do not directly target drivelological social-media videos, where the central meaning may be carried by a single decisive modality, such as on-screen wordplay or speech, or by non-linear interactions among captions, speech, visual action, editing, sound, and situated cultural context. Thus, our benchmark treats multimodality as the interpretive setting, not as a requirement that every example use all modalities.




Closer to our setting, MUStARD \cite{castro-etal-2019-towards} and MUStReason \cite{saha2025mustreason} study sarcasm in audiovisual dialogue, while UR-FUNNY \cite{hasan2019ur}, SMILE \cite{hyun-etal-2024-smile}, v-HUB \cite{shi2026vhubbenchmarkvideohumor}, and D-HUMOR \cite{kasu2025d} evaluate humour across speech and video. SIV-Bench \cite{kong2025siv} focuses on social interaction reasoning, and ViMU \cite{li2026vimubenchmarkingvideometaphorical} evaluates video metaphorical understanding through interpretation, rhetorical mechanism identification, social value identification, and evidence grounding. 
As summarised in Table~\ref{tab:benchmark_comparison}, these benchmarks cover important aspects of non-literal and socially grounded multimodal understanding, but typically focus on one phenomenon, one domain, or one evaluation format. 

Drivelology was introduced by \citet{wang2025drivel} to describe expressions that appear trivial, absurd, or nonsensical on the surface but are intentionally structured to convey an implicit humour, satire, or stance. This distinguishes drivelological content from ordinary nonsense or vacuous language \cite{frankfurt2009bullshit, cappelen2019bad}: its apparent absurdity is not merely empty, but functions as a cue for pragmatic interpretation. The original DrivelHub benchmark studies this phenomenon in text, showing that LLMs often misread purposeful absurdity as shallow nonsense and fail to recover the underlying narrative \cite{wang2025drivel}.

Our work examines the same problem in short-form video, where the cue that makes an apparently nonsensical expression meaningful may appear in speech, on-screen text, visual action, editing, audio, cultural reference, or their interaction. This setting introduces two additional challenges beyond text. First, the relevant evidence may be distributed unevenly across modalities: some examples depend mainly on a single modality, such as wordplay in captions, while others require integrating incongruent cues across modalities. Second, the implicit meaning often depends on temporal presentation, such as a delayed reveal, a cut, or a change in framing. The resulting task is therefore not only to recognise non-literal language, but to explain how multimodal evidence supports a particular pragmatic interpretation. 
We also connect drivelology evaluation to representation-level approaches to reasoning. Standard open-ended or multiple-choice evaluations can conflate perception, reasoning, instruction following, decoding behaviour, and verbosity \cite{loakman-etal-2025-comparing}. Reasoning-as-retrieval \cite{xiao2024rar} provides a complementary diagnostic by asking whether a query is close to the correct answer in representation space. We adapt this idea to bidirectional video-text retrieval: given a video, a model retrieves its implicit narrative, and given a narrative, it retrieves the corresponding video. This does not replace explanation-based evaluation; rather, it tests whether videos and their implicit interpretations are aligned in the model's representation space, beyond surface visual or lexical similarity.

\section{Multimodal Drivelology Benchmark}
\label{sec:benchmark}

\subsection{Task Definition}
\label{sec:task-definition}

We define a \textit{drivelological video} as a short-form video whose observable content is interpretable at the surface level, but whose implicit meaning is not fully captured by a literal description of what is seen or heard. The surface layer may be mundane or seemingly nonsensical, but it is constructed so that viewers can infer an implicit pragmatic meaning, such as humour, satire, stance, ironic reversal, or affective implication. 
In the multimodal setting, the implicit interpretation may be supported by a single decisive modality or by interactions among modalities. For example, an on-screen caption may contain the key wordplay, speech may deliver the punchline, or a visual transition may create the humour. In other cases, the meaning depends on how modalities are paired: the same text, utterance, or visual action may imply different meanings when combined with congruent or incongruent images, actions, sounds, editing, or cultural context. Thus, the task is not simply to detect whether a video is humorous, ironic, or sarcastic, but to explain how its observable cues support a particular implicit interpretation.


Each DrivelHub+ instance consists of a video, a human-written implicit meaning explanation, and metadata indicating which modality or combination of modalities supports the interpretation. The goal is to evaluate whether a model can deliver the implicit meaning of the video, rather than merely describe its visible or audible content. The benchmark supports two complementary evaluation settings: explanation and representation-level bidirectional retrieval, which are described in Section~\ref{sec:experiments}.

\subsection{Data Collection and Filtering}
\label{sec:data-collection}


To collect video candidates for implicit multimodal reasoning, we curated short-form videos from public social-media platforms, including Instagram, Threads, Facebook, YouTube, and TikTok. We prioritised cases in which a literal description of actions, captions, or speech would be insufficient to explain why the video is meaningful, humorous, ironic, satire, or socially pointed. 
A candidate video was retained only if it satisfied three criteria: (1) It contains an observable surface layer, such as a visible event, spoken utterance, on-screen caption, edited sequence, or staged presentation, that can be described literally. (2) It conveys an implicit pragmatic meaning beyond this literal description, such as humour, satire, stance, social implication, or ironic reversal. (3) The implicit meaning depends on a drivelological mechanism, such as misdirection, switchbait, inversion, exaggeration, wordplay, cultural reference, or a non-linear interaction among modalities.

We excluded videos that were purely random, lacked a recoverable implicit meaning, depended on inaccessible private context, or could be fully understood through surface-level description alone. We also removed examples containing harassment, hate speech, private personal information, or material likely to cause direct harm. The final  \textit{DrivelHub+} contains 1,000 social-media videos paired with human-written implicit meaning explanations.

\subsection{Annotation and Quality Control}
\label{sec:annotation-main}


All annotations in DrivelHub+ were produced by human annotators. No Video LLM or other generative model was used to generate the final explanations. The annotation team consisted of five annotators with experience in social-media. For each video, annotators watched the full clip and wrote a concise explanation of its implicit pragmatic meaning, such as the humour, satire, stance, affective implication, cultural reference, or rhetorical reversal it communicates. The target annotation was therefore an interpretation of the video's underlying point, rather than a scene description or transcript.

Annotators also provided structured metadata, including the modalities required for interpretation, the languages of speech and on-screen text, and sensitive-content information for transparency and filtering. The modality label records which streams the interpretation depends on rather than which streams the clip happens to contain: background music that could be muted without changing the reading does not make a clip audio-dependent, whereas a tone of voice that reverses the meaning of an otherwise ordinary visual event does. Appendix~\ref{app:modality-annotation} gives the definition and the distribution, and Appendix~\ref{app:cascade} uses these labels to group clips. Sensitive-content labels indicate whether a video contains potentially sensitive material, such as racist content, sexual content, or dark humour, and should not be interpreted as endorsement. 

Every annotator worked through the whole corpus rather than a share of it. At a sustained rate of about five videos per hour, one pass over the 1,000 clips takes roughly 200 hours, so the five independent passes come to about 1,000 hours of annotation before any adjudication. The five readings of each clip were then compared in a consensus discussion, which decided whether to keep one of them, merge them, revise the explanation, or drop the clip altogether when no recoverable drivelological meaning could be agreed on. Candidate annotations were checked at this stage for being video-grounded pragmatic interpretations rather than literal descriptions. Two third-party annotators, who wrote none of the explanations, then reviewed all 1,000 retained clips to assess the consistency of the inclusion decisions, reaching $89\%$ agreement.



\subsection{Dataset Statistics}
\label{sec:dataset-statistics}

Table~\ref{tab:dataset-language-full} reports the distributions of speech language and caption language in DrivelHub+. 
Language labels indicate the linguistic signal used for interpretation: when speech contributes to meaning, the speech-language label records the spoken languages; when on-screen text contributes to meaning, the caption-language label records the relevant caption languages. If neither speech nor on-screen text is present, the corresponding language field is marked as \textit{None}. Since the dataset is predominantly English, these labels are intended primarily for transparency rather than balanced multilingual evaluation.

\section{Experiments}
\label{sec:experiments}

\subsection{Experimental Setup}
\label{sec:experimental-setup}

We evaluate model performance on DrivelHub+ in both explanation and retrieval settings. 
Both settings require the soundtrack, so we restrict the evaluation to open models that accept an audio stream alongside video. For the explanation setting these are Qwen3-Omni and Qwen2.5-Omni~\cite{xu2025qwen25omnitechnicalreport,xu2025qwen3omnitechnicalreport}, MiniCPM-o-2.6~\cite{yao2024minicpmvgpt4vlevelmllm}, Gemma 4~\cite{gemmateam2026gemma4}, and EchoInk-R1~\cite{xing2025echoinkr1exploringaudiovisualreasoning}. For retrieval, we compare embedding-native omni-modal models, including omni-embed-nemotron~\cite{xu2025omni}, LCO-Omni~\cite{xiao2026scaling}, e5-omni~\cite{chen2026e5omniexplicitcrossmodalalignment}, and jina-v5-omni~\cite{honicke2026jina}, against generative Video LLMs adapted for similarity-based retrieval. 
Full checkpoints and decoding settings are listed in Appendix~\ref{app:model-details}. 
We exclude proprietary closed-source models because their inference pipelines may involve undisclosed tool use, retrieval augmentation, or changing backend behaviour, making controlled evaluation difficult. 

\begin{table*}[t]
\centering
\resizebox{\linewidth}{!}{
\begin{tabular}{lllccccccccc}
\toprule
\textbf{Model} 
& \textbf{Size}
& \textbf{Mode}
& \textbf{Aligned\%} $\uparrow$
& \textbf{Core/5} $\uparrow$
& \textbf{Rhet./3} $\uparrow$
& \textbf{Social/2} $\uparrow$
& \textbf{Ground./2} $\uparrow$
& \textbf{Halluc./3} $\downarrow$
& \textbf{Literal/3} $\downarrow$
& \textbf{Vague/2} $\downarrow$
& \textbf{Total/12} $\uparrow$ \\
\midrule
\textcolor{gray}{Human baseline} & \textcolor{gray}{--} & \textcolor{gray}{--}
& \textcolor{gray}{0.997} & \textcolor{gray}{3.954} & \textcolor{gray}{1.997} & \textcolor{gray}{1.008} & \textcolor{gray}{1.065} & \textcolor{gray}{0.007} & \textcolor{gray}{0.013} & \textcolor{gray}{0.008} & \textcolor{gray}{7.996} \\
\midrule

Qwen3-Omni & 30B-A3B & Thinking
& \bestA{0.614} & \bestA{2.888} & \bestA{1.635} & \bestB{0.942} & \bestA{1.039} & \bestA{0.390} & \bestA{0.361} & \bestA{0.396} & \bestA{5.357} \\
Qwen3-Omni & 30B-A3B & No-thinking
& \bestB{0.565} & \bestB{2.765} & \bestB{1.592} & \bestA{0.950} & \bestB{1.026} & \bestB{0.505} & \bestB{0.412} & \bestB{0.440} & \bestB{4.976} \\
EchoInk-R1 & 7B & Instruct
& \bestC{0.328} & \bestC{2.217} & \bestC{1.300} & \bestC{0.909} & 1.006 & \bestC{0.516} & 1.040 & 0.766 & \bestC{3.110} \\
Qwen2.5-Omni & 7B & Instruct
& 0.304 & 2.188 & 1.265 & 0.904 & \bestC{1.008} & 0.545 & 1.046 & \bestC{0.753} & 3.021 \\
Gemma 4 & E4B & No-thinking
& 0.187 & 1.912 & 1.153 & 0.860 & \worstB{0.994} & \worstB{0.719} & \bestC{0.996} & \worstB{0.896} & 2.308 \\
Gemma 4 & E2B & No-thinking
& \worstB{0.102} & \worstB{1.726} & \worstB{1.040} & \worstC{0.858} & \worstA{0.985} & \worstA{0.756} & \worstC{1.162} & \worstA{1.003} & \worstC{1.688} \\
Qwen2.5-Omni & 3B & Instruct
& \worstC{0.156} & \worstC{1.809} & \worstC{1.076} & \worstB{0.838} & \worstC{0.997} & \worstC{0.642} & \worstB{1.610} & \worstC{0.829} & \worstB{1.639} \\
MiniCPM-o-2.6 & 9B & Instruct
& \worstA{0.061} & \worstA{1.386} & \worstA{0.914} & \worstA{0.717} & 1.004 & 0.517 & \worstA{1.959} & 0.807 & \worstA{0.738} \\
\midrule
\multicolumn{12}{l}{\textit{\textcolor{gray}{Text-only cascade: the clip converted to text, no video}}} \\
\textcolor{gray}{OCR + ASR} & \textcolor{gray}{--} & \textcolor{gray}{--}
& \textcolor{gray}{0.441} & \textcolor{gray}{2.443} & \textcolor{gray}{1.446} & \textcolor{gray}{0.920} & \textcolor{gray}{1.014} & \textcolor{gray}{0.668} & \textcolor{gray}{0.571} & \textcolor{gray}{0.580} & \textcolor{gray}{4.004} \\
\textcolor{gray}{ASR only} & \textcolor{gray}{--} & \textcolor{gray}{--}
& \textcolor{gray}{0.345} & \textcolor{gray}{2.153} & \textcolor{gray}{1.325} & \textcolor{gray}{0.897} & \textcolor{gray}{0.984} & \textcolor{gray}{0.866} & \textcolor{gray}{0.675} & \textcolor{gray}{0.668} & \textcolor{gray}{3.150} \\
\textcolor{gray}{OCR only} & \textcolor{gray}{--} & \textcolor{gray}{--}
& \textcolor{gray}{0.202} & \textcolor{gray}{1.853} & \textcolor{gray}{1.188} & \textcolor{gray}{0.867} & \textcolor{gray}{0.981} & \textcolor{gray}{0.975} & \textcolor{gray}{0.924} & \textcolor{gray}{0.819} & \textcolor{gray}{2.171} \\

\bottomrule
\end{tabular}
}
\caption{
Video LLM-as-a-judge results for implicit-meaning understanding on 1,000 drivelological videos, graded by gemini-3.8-flash. Green shading marks the top three results in each column and purple shading the bottom three, computed over the models only; the human baseline is the reference ceiling and is excluded from the shading. The cascade rows convert each clip to text with off-the-shelf speech and character recognition and give a text-only model nothing else; they are diagnostics on the same scale rather than competing systems, and are also excluded from the shading.
}
\label{tab:generation}
\end{table*}

\begin{table*}[t]
\centering
\small
\resizebox{\linewidth}{!}{
\begin{tabular}{lcccccccccccc}
\toprule
\multirow{2}{*}{\textbf{Model}}
& \multicolumn{6}{c}{\textbf{Text-to-Video}}
& \multicolumn{6}{c}{\textbf{Video-to-Text}} \\
\cmidrule(lr){2-7} \cmidrule(lr){8-13}
& R@1 & R@5 & R@10 & MRR & N@5 & N@10
& R@1 & R@5 & R@10 & MRR & N@5 & N@10 \\
\midrule

\multicolumn{13}{l}{\textit{Embedding-native retrieval models}} \\

omni-embed-nemotron-3b
& \bestA{87.4} & \bestA{95.5} & \bestA{97.2} & \bestA{92.2} & \bestA{92.7} & \bestA{93.2}
& \bestA{87.7} & \bestA{96.2} & \bestA{96.8} & \bestA{92.5} & \bestA{93.2} & \bestA{93.4} \\

LCO-Omni-3B
& \bestC{82.2} & \bestC{93.0} & \bestC{95.2} & \bestC{88.2} & \bestC{88.8} & \bestC{89.6}
& \bestB{80.5} & \bestB{91.7} & \bestB{94.7} & \bestB{86.7} & \bestB{87.4} & \bestB{88.3} \\

LCO-Omni-7B
& \bestB{83.5} & \bestB{93.7} & \bestB{95.8} & \bestB{89.4} & \bestB{89.9} & \bestB{90.6}
& \bestC{79.1} & \bestC{91.2} & \bestC{93.7} & \bestC{85.7} & \bestC{86.4} & \bestC{87.3} \\

jina-v5-omni-nano
& 54.5 & 73.1 & 78.6 & 63.8 & 64.9 & 66.7
& 52.2 & 70.8 & 76.5 & 61.4 & 62.5 & 64.4 \\

jina-v5-omni-small
& 45.5 & 63.8 & 72.3 & 55.6 & 55.8 & 58.6
& 39.6 & 59.3 & 66.3 & 49.6 & 50.5 & 52.8 \\

e5-omni-7B
& 66.4 & 81.3 & 86.3 & 74.2 & 74.9 & 76.5
& 68.5 & 83.0 & 86.1 & 76.4 & 77.1 & 78.1 \\

\midrule

\multicolumn{13}{l}{\textit{Representative generative models adapted for retrieval}} \\

Qwen2.5-Omni-3B
& \worstC{21.2} & \worstC{32.5} & \worstC{39.5} & \worstC{27.9} & \worstC{27.3} & \worstC{29.6}
& \worstC{19.3} & \worstC{30.5} & \worstC{35.2} & \worstC{25.7} & \worstC{25.4} & \worstC{26.9} \\

Qwen2.5-Omni-7B
& \worstB{18.2} & \worstB{30.4} & \worstB{37.4} & \worstB{25.2} & \worstB{24.9} & \worstB{27.1}
& \worstB{16.6} & \worstB{27.8} & \worstB{34.5} & \worstB{23.2} & \worstB{22.7} & \worstB{24.9} \\

Qwen3-Omni-30B-A3B
& \worstA{0.5} & \worstA{0.9} & \worstA{1.8} & \worstA{1.4} & \worstA{0.7} & \worstA{1.0}
& \worstA{0.1} & \worstA{0.8} & \worstA{1.2} & \worstA{1.0} & \worstA{0.5} & \worstA{0.6} \\

\bottomrule
\end{tabular}
}
\caption{Text-to-video and video-to-text retrieval performance on the revised DrivelHub+ split. Scores are percentages. Green shading marks the top three results in each column, and purple shading marks the bottom three.}
\label{tab:retrieval}
\end{table*}

\subsection{Explanation Evaluation}
\label{sec:generation-eval-main}

We use a structured Video LLM-as-a-judge protocol that compares generated explanations against human annotations and reports semantic alignment and a rubric-based total score. 
The rubric measures whether a response captures the main implicit point of the video, identifies the rhetorical mechanism that makes it work, preserves the relevant social or affective implication, and grounds the interpretation in the appropriate multimodal evidence. It also penalises hallucinated, literal-only, and overly vague responses. 

We use Gemini 3.8 Flash \cite{gemini38flash2026} as the judge for all systems. The judge must receive the same video \emph{and} soundtrack as the systems it grades, which rules out judges without an audio encoder, and it should not belong to any evaluated family. 
Using a proprietary model as the judge is consistent with excluding proprietary models from the evaluation itself. 
Because a hosted model cannot be rerun by readers without cost, we also graded every explanation with an open audio-capable judge, Qwen3-Omni-30B-A3B-Instruct. The two gradings agree on the ordering\footnote{Spearman $\rho = 0.70$, Pearson $r = 0.80$, and per-explanation Cohen's $\kappa = 0.44$. Appendix~\ref{app:judge-robustness} reports both gradings and the one system on which they disagree.} and place the same three systems at the top, so the open judge reproduces our conclusions without an API key. We discuss the full scoring rubric in Appendix~\ref{app:judge-protocol}. Appendix~\ref{app:cascade} additionally reports a cascaded baseline, which separates failures of modality extraction from failures of pragmatic inference. Appendix~\ref{app:vision-gap} then shows that every evaluated system loses accuracy on the clips whose meaning is carried by vision, while a human annotator does not, so that subset is hard for the models rather than hard in itself.

\subsection{Representation-Level Retrieval}
\label{sec:retrieval-eval-main}




We adapt reasoning-as-retrieval \cite{xiao2024rar} as bidirectional video-text retrieval, testing whether videos and their implicit narratives are aligned in representation space. Candidates are ranked by representation similarity. Retrieval relevance is defined using annotated relevance judgments: the original video-narrative pair is always treated as relevant, and additional relevant pairs are included when different samples share highly consistent implicit meanings. 

To identify these additional pairs, we first embed all narrative texts using EmbeddingGemma-300M \cite{vera2025embeddinggemmapowerfullightweighttext} and retrieve the top-10 nearest narrative candidates for each of the 1,000 samples. 
Candidate pairs with cosine similarity above 0.7 are then manually reviewed to determine whether the two narratives express the same or highly consistent implicit meaning. The resulting human-verified matches are added as additional relevance annotations. For models requiring instruction-formatted embeddings, prompts are given in Appendix~\ref{app:retrieval-prompt}. We report Recall@K, MRR, and NDCG@K for both retrieval directions.



\begin{figure*}[t]
    \centering
    \includegraphics[width=\textwidth]{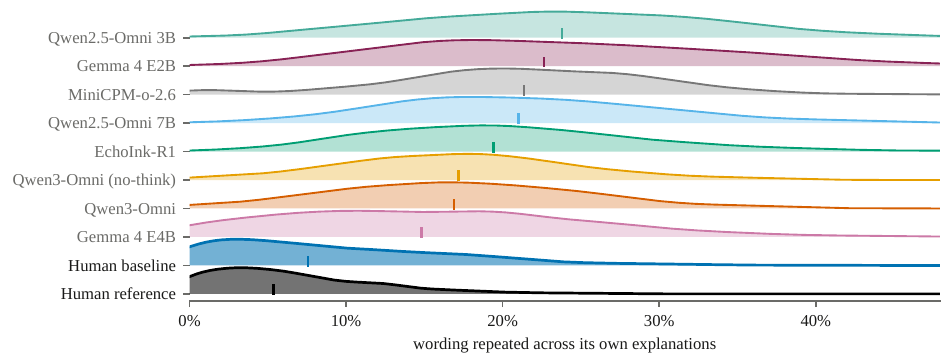}
    \caption{%
        Every row wrote an explanation for each of the same $877$ clips. For one of those explanations we take its three-word phrases and ask how many of them also appear in that same row's explanations of \emph{other} clips; the curve is the distribution of that percentage over all its explanations, and the tick mark is the median. A row that opens many answers with the same formula sits to the right; a row that words each answer to its own clip sits to the left. Appendix~\ref{app:style} gives the procedure.
    }
    \label{fig:rarity}
\end{figure*}

\section{Results and Discussion}
\label{sec:results-discussion}

\paragraph{Implicit meaning explanation.}
Table~\ref{tab:generation} reports results for the free-form implicit meaning explanation setting. 
The best-performing model is Qwen3-Omni-30B-A3B in thinking mode, with an alignment rate of $61.4\%$ and a total score of $5.36$ out of $12$, against $8.00$ for the human reference: the strongest system reaches roughly two thirds of the human score, and barely half of its answers are judged aligned at all. The margin over its own non-thinking mode is small ($5.36$ versus $4.98$), so test-time reasoning is not by itself what separates the models here. The judge scores the human reference at $8.00$ rather than near the $12$-point ceiling, mostly on the social-meaning and grounding dimensions; absolute totals should therefore be read as positions on a strict scale rather than as percentages of achievable understanding. 

Appendix~\ref{app:reasoning-trace-case-study} works through one clip on which all eight systems are judged unaligned while a human one-sentence answer scores $8$ of $12$. The lesson generalises: useful reasoning has to preserve the frame-level transition that carries the implicit meaning, and producing more detail about the scene does not substitute for it.

\paragraph{Retrieval task.}
Explanation probes what a model can produce under instruction; retrieval probes what its representation already encodes, whether a video and its written interpretation sit near one another in a shared space with no generation involved. Comparing an embedding-native retriever against a generative Video LLM adapted for retrieval is therefore not a fair ranking: one was trained for this objective and the other was not. What Table~\ref{tab:retrieval} supports is the narrower claim that the training objective, rather than scale or the generative ability of the backbone, determines whether drivelological meaning is recoverable from the representation. 
The controlled comparison inside the table carries that claim. LCO-Omni is trained from Qwen2.5-Omni, so the pair differs in objective while sharing a backbone: LCO-Omni-7B reaches $83.5$ R@1 text-to-video against Qwen2.5-Omni-7B's $18.2$, and LCO-Omni-3B $82.2$ against $21.2$. The generative backbone already sits far above the $0.1\%$ of a random pick among the 1,000 candidates, so the pragmatic signal is present in its representation; contrastive training on the same weights raises it fourfold, which suggests the signal was largely inaccessible to nearest-neighbour search until the objective reorganised the space. This is consistent with the view that representation quality remains bounded by the generative capacity of the base model \cite{xiao2026scaling}: contrastive training improves alignment but cannot create robust pragmatic representations from a weak backbone.
The two retrieval directions are also not interchangeable: Appendix~\ref{app:retrieval-direction-cases} shows clips that the same embeddings place at rank $1$ one way and past rank $150$ the other, which is why Table~\ref{tab:retrieval} reports them separately rather than averaged. Retrieval depends on relative ranking rather than absolute similarity, and Appendix~\ref{app:score-space} shows that a high cosine value alone is not evidence of a correct match.


Among the embedding-native systems, where the comparison is a fair one, omni-embed-nemotron-3b leads every column of Table~\ref{tab:retrieval}, at $87.4$ R@1 text-to-video and $87.7$ video-to-text against $83.5$ and $80.5$ for the strongest LCO-Omni variant in each direction. Neither scale nor the embedding interface accounts for this: it has $3$B parameters and a 2,048-dimensional vector, both smaller than LCO-Omni-7B, which it beats everywhere, while jina-v5-omni is also a native encoder and is last in this group. The table establishes the ranking but not its cause, which lies in training data and objectives we do not isolate. The consequence for the benchmark is that headroom on the retrieval task is smaller than the generative results suggest: a system trained for video-text alignment already recovers most of it.

\begin{figure*}[tp]
    \centering
    \includegraphics[width=\textwidth]{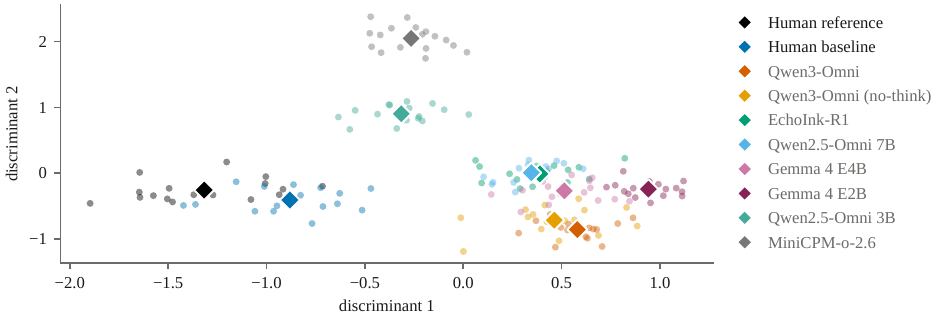}
    \caption{%
        Who wrote an explanation is recoverable from its style alone. Every content word is discarded here: each point is a group of $25$ explanations placed only by how often its writer reaches for function words and punctuation, which say nothing about what the clip was about. 
        Diamonds mark each writer's centre. Appendix~\ref{app:style} gives the construction; the analysis follows \citet{russell2026storyscope}.
    }
    \label{fig:lda}
\end{figure*}

\paragraph{Retrieval directions.}
Retrieval is not a symmetric measure of a single ability. Text-to-video asks whether a written interpretation can single out one clip among many similar ones; video-to-text asks whether the clip can be encoded close to that interpretation. Most systems lose accuracy in the second direction, the two LCO-Omni variants swap places between the two, and individual clips can sit at rank $1$ one way and past rank $150$ the other. We therefore report both directions rather than averaging them; Appendix~\ref{app:retrieval-direction-cases} gives the per-query evidence.

\paragraph{How the explanations are written.}
The rubric asks whether an explanation recovers the meaning, not how it is written, so the scores say nothing about style. Two measures over the same texts do, and both separate the human annotations from every model. Both count word tokens, so both are restricted to the $877$ clips whose human annotation is written in the Latin alphabet. Every system explained the same clips, so each produced one explanation per clip. Figure~\ref{fig:rarity} asks how often a system repeats itself across that set: what share of one explanation's three-word phrases also turn up in its explanations of other clips. The human annotations repeat $5.4\%$ and $7.6\%$ of theirs; every model repeats between $14.8\%$ and $23.8\%$, three to four times as much. The models keep returning to the same formulations from one video to the next, while the annotators word each explanation to its own clip. That could still be a quirk of one measure, so Figure~\ref{fig:lda} asks whether the difference survives throwing away everything the explanations are about. Keeping only how often a writer reaches for function words and punctuation, and discarding every content word, still separates the writers, with the annotators apart from all eight models; a classifier given nothing but those rates tells a human explanation from a model one at macro-$F_1$ $0.81$. The vocabulary that separates them is interpretable: models reach for \textit{video}, \textit{scene}, \textit{screen}, \textit{person} and \textit{character}, humans for \textit{because}, \textit{literally}, \textit{supposedly}, \textit{instead} and \textit{deliberately}. Models narrate what is displayed; the human writes the inference that closes the gap. A benchmark scored only on meaning recovery would record none of this.


\section{Conclusion}
\label{sec:conclusion}

We introduced \textit{DrivelHub+}, a benchmark for Multimodal Drivelology, to evaluate implicit pragmatic meaning in short-form social-media videos. The benchmark contains 1,000 videos with human-written implicit-meaning explanations, targeting cases where literal descriptions of actions, speech, or captions are insufficient. 
We evaluated models through free-form explanation and bidirectional retrieval. Results show that current Video LLMs still struggle with pragmatic multimodal comprehension, especially when meaning depends on subtle visual cues, wordplay, cultural knowledge, or cross-modal interaction. Thinking models achieve the strongest generation results, but only when reasoning remains grounded in decisive visual evidence. Retrieval results further show that generation ability and representation alignment are not interchangeable. 
Overall, DrivelHub+ highlights a persistent gap between surface-level multimodal recognition and pragmatic interpretation: models can often describe what is shown, but still fail to infer what is meant.

\section*{Limitations}
\label{sec:limitations}

DrivelHub+ focuses on culturally situated implicit meaning, which is inherently interpretive; this is a property of the phenomenon under study rather than an artefact of dataset construction. The benchmark is predominantly English by design, and balanced multilingual generalisation is outside the scope of this release. To support reproducibility while accounting for sensitive content, the videos, annotations, and metadata are released on HuggingFace under a gated research-use license; access requires an application and is restricted to research purposes. 

\section*{Disclosure}

We used LLMs to assist with drafting and restructuring parts of the main text and appendices. The authors reviewed and revised the resulting material and take responsibility for the methods, experimental results, claims, and references in the final manuscript.

\section*{Ethics Statement}
\label{sec:ethics}

DrivelHub+ is constructed from publicly available social-media videos from Instagram, Threads, Facebook, YouTube, and TikTok. 
The study was approved by the authors' institutional ethics review board. The videos were collected manually by viewing public posts and saving them one at a time, without automated scraping. 
We do not collect private messages, restricted-access content, or unnecessary user metadata such as usernames, profile information, comments, or follower counts, and the dataset is intended only for academic research on multimodal pragmatic understanding. Because drivelological videos may involve humour, satire, stereotypes, social criticism, sexual references, dark humour, or culturally sensitive content, some examples may contain offensive or sensitive material; such content is included only for analysis, does not reflect the authors' views, and was reviewed to remove harassment, hate speech, private personal information, or material likely to cause direct harm. 
These metadata categories are descriptive and non-exhaustive. 
Annotators were informed about possible sensitive content and could skip or flag difficult, ambiguous, or uncomfortable examples. We document sensitive examples with metadata where appropriate and recommend that the benchmark not be used for surveillance, profiling, moderation decisions, or decision-making about individuals or communities. 
We release the DrivelHub+ benchmark, including videos, annotations, metadata, prompts, and evaluation scripts, on HuggingFace under a gated research-use license. Access is granted only for academic research purposes and is subject to the license terms. The original videos remain subject to the platforms' terms of service and the rights of their creators. Creators of a video, or people who appear in it, can request through the dataset page that the video and its annotations be removed, and we will remove them from the released data.


\bibliography{custom}

\appendix

\clearpage

\section{Model Details}
\label{app:model-details}

In this work, we evaluate and compare several state-of-the-art Video LLMs and multimodal embedding models. 
For generation, we include Qwen3-Omni \cite{xu2025qwen3omnitechnicalreport}, Qwen2.5-Omni \cite{xu2025qwen25omnitechnicalreport}, MiniCPM-o-2.6 \cite{yao2024minicpmvgpt4vlevelmllm}, Gemma 4 \cite{gemmateam2026gemma4}, and EchoInk-R1 \cite{xing2025echoinkr1exploringaudiovisualreasoning}; every one of them accepts an audio stream, which is a precondition for this benchmark. For retrieval, we include embedding-native models such as omni-embed-nemotron \cite{xu2025omni}, LCO-Embedding-Omni \cite{xiao2026scaling}, e5-omni \cite{chen2026e5omniexplicitcrossmodalalignment}, and jina-v5-omni \cite{honicke2026jina}, together with representative generative Video LLMs adapted for similarity-based retrieval.

Table~\ref{tab:model-stats} reports the exact checkpoints and sampling parameters used for generation. When official inference recommendations are available, we follow them. Otherwise, we use the decoding parameters listed in the table. For models with both thinking and non-thinking modes, we evaluate the corresponding modes separately. All generation experiments were served with vLLM \cite{kwon2023efficient} on nodes equipped with four NVIDIA GH200 Grace Hopper Superchips. Table~\ref{tab:retrieval-model-stats} lists the retrieval checkpoints, the dimensionality of each embedding, and how the vector is obtained.

\begin{table*}[t]
\centering
\small
\resizebox{\linewidth}{!}{
\begin{tabular}{llccccc}
\toprule
\textbf{Model} & \textbf{Mode} & \textbf{Temp.} & \textbf{Top-P} & \textbf{Top-K} & \textbf{Context} & \textbf{Checkpoint} \\
\midrule

\multirow{2}{*}{Qwen3-Omni-30B-A3B}
& thinking
& 0.6 & 0.95 & - & 65,536
& \multirow{2}{*}{\href{https://huggingface.co/Qwen/Qwen3-Omni-30B-A3B-Thinking}{Qwen/Qwen3-Omni-30B-A3B-Thinking}} \\

& non-thinking
& 0.7 & 0.80 & - & 65,536
& \\
\midrule

MiniCPM-o-2.6-9B
& -
& 0.7 & 0.70 & - & 32,768
& \href{https://huggingface.co/openbmb/MiniCPM-o-2_6}{openbmb/MiniCPM-o-2\_6} \\
\midrule

EchoInk-R1-7B
& -
& 0.7 & 0.95 & - & 32,768
& \href{https://huggingface.co/harryhsing/EchoInk-R1-7B}{harryhsing/EchoInk-R1-7B} \\
\midrule

Qwen2.5-Omni-7B
& -
& 0.7 & 0.80 & - & 32,768
& \href{https://huggingface.co/Qwen/Qwen2.5-Omni-7B}{Qwen/Qwen2.5-Omni-7B} \\

Qwen2.5-Omni-3B
& -
& 0.7 & 0.80 & - & 32,768
& \href{https://huggingface.co/Qwen/Qwen2.5-Omni-3B}{Qwen/Qwen2.5-Omni-3B} \\
\midrule

Gemma-4-E4B-it
& non-thinking
& 1.0 & 0.95 & 64 & 131,072
& \href{https://huggingface.co/google/gemma-4-E4B-it}{gemma-4-E4B-it} \\

Gemma-4-E2B-it
& non-thinking
& 1.0 & 0.95 & 64 & 131,072
& \href{https://huggingface.co/google/gemma-4-E2B-it}{google/gemma-4-E2B-it} \\
\midrule

\multicolumn{7}{l}{\textit{Video LLM Judge}} \\

Gemini 3.8 Flash
& -
& - & - & - & -
& - \\

Qwen3-Omni-30B-A3B-Instruct
& -
& 0.7 & 0.80 & - & 65,536
& \href{https://huggingface.co/Qwen/Qwen3-Omni-30B-A3B-Instruct}{Qwen/Qwen3-Omni-30B-A3B-Instruct} \\

\bottomrule
\end{tabular}
}
\caption{Model checkpoints and sampling parameters for generation used in experiments. Every model listed accepts an audio stream, and every run passed the clip's soundtrack alongside its frames. A dash under Top-K means the flag was left unset, so the checkpoint's own generation configuration applies. All models were served with vLLM in bfloat16, with the context length shown; for MiniCPM-o-2.6 the soundtrack is split into 30-second chunks, because vLLM caps that model at 30 one-second audio chunks per request.}
\label{tab:model-stats}
\end{table*}

\begin{table*}[t]
\centering
\small
\begin{tabular}{llll}
\toprule
\textbf{Model} & \textbf{Dim.} & \textbf{Embedding} & \textbf{Checkpoint} \\
\midrule

\multicolumn{4}{l}{\textit{Embedding-native retrieval models}} \\

omni-embed-nemotron-3b & 2,048 & Native encoder
& \href{https://huggingface.co/nvidia/omni-embed-nemotron-3b}{nvidia/omni-embed-nemotron-3b} \\

LCO-Omni-7B & 3,584 & Last-token hidden state
& \href{https://huggingface.co/LCO-Embedding/LCO-Embedding-Omni-7B}{LCO-Embedding/LCO-Embedding-Omni-7B} \\

LCO-Omni-3B & 2,048 & Last-token hidden state
& \href{https://huggingface.co/LCO-Embedding/LCO-Embedding-Omni-3B}{LCO-Embedding/LCO-Embedding-Omni-3B} \\

e5-omni-7B & 3,584 & Last-token hidden state
& \href{https://huggingface.co/Haon-Chen/e5-omni-7B}{Haon-Chen/e5-omni-7B} \\

jina-v5-omni-nano & 768 & Native encoder
& \href{https://huggingface.co/jinaai/jina-embeddings-v5-omni-nano-retrieval}{jinaai/jina-embeddings-v5-omni-nano-retrieval} \\

jina-v5-omni-small & 1,024 & Native encoder
& \href{https://huggingface.co/jinaai/jina-embeddings-v5-omni-small-retrieval}{jinaai/jina-embeddings-v5-omni-small-retrieval} \\

\midrule

\multicolumn{4}{l}{\textit{Generative models adapted for retrieval}} \\

Qwen2.5-Omni-3B & 2,048 & Last-token hidden state
& \href{https://huggingface.co/Qwen/Qwen2.5-Omni-3B}{Qwen/Qwen2.5-Omni-3B} \\

Qwen2.5-Omni-7B & 3,584 & Last-token hidden state
& \href{https://huggingface.co/Qwen/Qwen2.5-Omni-7B}{Qwen/Qwen2.5-Omni-7B} \\

Qwen3-Omni-30B-A3B & 2,048 & Last-token hidden state
& \href{https://huggingface.co/}{Qwen/Qwen3-Omni-30B-A3B-Thinking} \\

\bottomrule
\end{tabular}
\caption{
Retrieval models, embedding dimensionality, and how each vector is obtained. Every input carries the clip's soundtrack alongside its frames. \textit{Native encoder} means the checkpoint's own embedding interface produces the vector; \textit{last-token hidden state} means the final hidden state is pooled under the embedding prompt of Appendix~\ref{app:retrieval-prompt}. For jina-v5-omni we use the retrieval-specific checkpoints. Vectors are L2-normalised before scoring, and retrieval uses cosine similarity.
}
\label{tab:retrieval-model-stats}
\end{table*}

\section{Dataset Statistics}
\label{app:dataset-statistics}

We provide additional dataset statistics beyond the summary reported in \S\ref{sec:dataset-statistics}. We include full language distributions and sensitive-content remark distributions to support transparency and responsible use of the benchmark.

\begin{table*}[t]
\centering
\resizebox{\linewidth}{!}{
\begin{tabular}{llllccc}
\toprule
\textbf{Benchmark} & \textbf{Domain} & \textbf{Focus} & \textbf{Output} & \textbf{Implicit?} & \textbf{Broad?} & \textbf{Social?} \\
\midrule
MUStARD \cite{castro-etal-2019-towards} & TV sitcoms & Sarcasm & Classification & Partial & No & No \\
MUStReason \cite{saha2025mustreason} & TV sitcoms & Sarcasm reasoning & Rationale + counterfactual & Yes & No & No \\
UR-FUNNY \cite{hasan2019ur} & TED Talks & Humour & Classification & Partial & No & No \\
SMILE \cite{hyun-etal-2024-smile} & TED/sitcoms & Laughter reasoning & Rationale & Yes & No & No \\
v-HUB \cite{shi2026vhubbenchmarkvideohumor} & Web videos & Visual humor & Classification + explanation + localisation & Yes & Limited & Partial \\
D-HUMOR \cite{kasu2025d} & Reddit memes & Dark humour & Classification + reasoning & Yes & Limited & Partial \\
SIV-Bench \cite{kong2025siv} & In-the-wild videos & Social interaction & Multiple-choice Question & Yes & Yes & Partial \\
ViMU \cite{li2026vimubenchmarkingvideometaphorical} & Creative videos & Metaphor & Multiple-choice Question + explanation & Yes & Limited & No \\
\midrule
DrivelHub+ (ours) & Short-form social media & Drivelology & Explanation + retrieval & Yes & Yes & Yes \\
\bottomrule
\end{tabular}
}
\caption{
Comparison with representative multimodal benchmarks for non-literal, humorous, and pragmatic video understanding.
\textbf{Implicit?} indicates whether the benchmark evaluates meanings beyond literal perception;
\textbf{Broad?} indicates whether it covers a wide range of pragmatic/rhetorical phenomena rather than a single category such as sarcasm, humor, or metaphor;
and \textbf{Social?} indicates whether the data is native to short-form or social-media-style communication.
}
\label{tab:benchmark_comparison}
\end{table*}

\subsection{Language Distributions}
\label{app:language-distributions}

Table~\ref{tab:dataset-language-full} reports the full distribution of speech-language and caption-language labels. We record language labels according to the linguistic signal used to infer the underlying meaning of each video. When speech is present, the speech-language label is determined by the spoken language or languages. When on-screen text contributes to interpretation, the caption-language label records the language or languages of the relevant text. If neither speech nor on-screen text is present, the corresponding language field is marked as \textit{None}. When multiple languages contribute to interpretation, all relevant languages are recorded.

\begin{table}[t]
\centering
\small
\setlength{\tabcolsep}{5pt}
\begin{tabular}{llr}
\toprule
\textbf{Category} & \textbf{Label} & \textbf{Count} \\
\midrule
\multirow{21}{*}{Speech}
& English & 757 \\
& \textit{None} & 138 \\
& Mandarin & 72 \\
& Cantonese & 4 \\
& Korean & 4 \\
& Taiwanese Hokkien & 4 \\
& English+Mandarin & 3 \\
& Japanese & 3 \\
& French & 2 \\
& Mandarin+Taiwanese Hokkien & 2 \\
& Arabic+English & 1 \\
& Cantonese+English & 1 \\
& Cantonese+Mandarin & 1 \\
& English+French+Spanish & 1 \\
& English+Icelandic & 1 \\
& English+Japanese & 1 \\
& English+Spanish & 1 \\
& Japanese+Mandarin & 1 \\
& Korean+Mandarin & 1 \\
& Spanish & 1 \\
& Thai & 1 \\

\midrule
\multirow{10}{*}{Caption}
& English & 780 \\
& Mandarin & 103 \\
& English+Mandarin & 68 \\
& \textit{None} & 40 \\
& Cantonese+Mandarin & 2 \\
& Japanese+Mandarin & 2 \\
& Korean+Mandarin & 2 \\
& English+Spanish & 1 \\
& French+Mandarin & 1 \\
& Japanese & 1 \\
\bottomrule
\end{tabular}
\caption{
Dataset distributions by speech language and caption language. For language metadata, multiple languages are recorded when multiple linguistic signals contribute to interpretation. The label \textit{None} indicates that no corresponding linguistic signal is present.
}
\label{tab:dataset-language-full}
\end{table}

\subsection{Sensitive Content Remarks}
\label{app:sensitive-content}

Because the dataset is collected from social-media videos, some examples contain potentially offensive, discriminatory, sexual, dark humour, or culturally sensitive material. Annotators record sensitive-content remarks to support transparent and responsible use of the dataset. These remarks are descriptive metadata only and do not imply endorsement of the underlying content. Table~\ref{tab:sensitive-content-remarks} reports the distribution of sensitive-content remark categories.

\begin{table}[t]
\centering
\small
\begin{tabular}{lr}
\toprule
\textbf{Category} & \textbf{Count} \\
\midrule
None & 858 \\
Racist & 54 \\
Sexual content & 47 \\
Dark humour & 41 \\
\bottomrule
\end{tabular}
\caption{
Distribution of sensitive-content remark categories in the dataset. The label \textit{None} denotes examples without an additional sensitive-content remark. These categories are included to document potentially sensitive material and support transparent use of the benchmark; they do not reflect the views of the authors.
}
\label{tab:sensitive-content-remarks}
\end{table}

\section{Data Processing Details}
\label{app:data-processing}

To ensure compatibility across video-processing backends, we standardise all video assets using the H.264 encoding format. Although AV1 provides stronger compression efficiency, it can cause decoding inconsistencies or empty-frame errors in common computer vision libraries such as OpenCV. Since several Video LLMs rely on such libraries during frame extraction, inconsistent video encoding can introduce evaluation noise unrelated to model capability. Standardising videos to H.264 reduces these input-processing failures and ensures that all evaluated models receive valid visual inputs. 
We avoid unnecessary modification of video content beyond format standardisation. Where possible, videos are processed in a way that preserves the visual, auditory, and textual cues needed for interpretation. This is important because Drivelological meaning may depend on fine-grained details such as timing, cuts, facial expressions, sound effects, or on-screen text.

\section{Annotation Protocol}
\label{app:annotation-protocol}

\subsection{Annotator Recruitment}
\label{app:annotator-recruitment}

All final DrivelHub+ annotations were produced by human annotators. No Video LLM or other generative model was used to generate the final explanations. The main annotation team consisted of five annotators with experience in multimodal social-media content and humour analysis. They watched the full videos, wrote concise implicit-narrative explanations, and provided structured metadata, including required modalities, speech and caption languages, and sensitive-content labels.

The main annotators were members of the research team or close project collaborators, rather than crowdworkers recruited through a public annotation platform. This choice was motivated by the task's reliance on implicit humour, sarcasm, cultural references, rhetorical reversals, cross-modal incongruity, and potentially sensitive social-media content, which require familiarity with the task definition and annotation guidelines. The annotators were compensated as part of their research or project work rather than being paid by the number of clips annotated.

We also recruited two third-party annotators for an independent quality-control check. These annotators wrote none of the explanations. They reviewed all 1,000 retained clips to assess the consistency of the inclusion decisions and the recoverability of the drivelological meaning, reaching $89\%$ agreement. The annotation itself was not divided between the main annotators: each of the five worked through the entire corpus, which at about five videos per hour is roughly 200 hours each and about 1,000 hours in total, and the five readings of a clip were then reconciled in consensus discussion. Ambiguous cases were discussed, revised, or removed when no recoverable drivelological meaning was found.


\subsection{Annotation Task}
\label{app:annotation-task}

Annotators were asked to inspect each DrivelHub+ short-form social-media video and provide an implicit narrative explanation describing the underlying meaning of the video. The intended annotation target was not a literal description of visible or audible content, but rather the inferred social, cultural, humorous, or pragmatic meaning conveyed by the video. To ensure consistency, annotators followed the annotation guideline shown in Figure~\ref{fig:annotation_guideline}. The guideline instructed annotators to focus on the core implicit meaning, use the video's primary language, avoid literal transcription or unsupported interpretation, and keep explanations concise. 
For each example, annotators also recorded metadata, including the modality or modalities needed to infer the underlying meaning, speech-language labels, caption-language labels, and sensitive-content remarks when applicable. Multiple modality and language labels could be assigned when more than one signal contributed to interpretation.

\begin{figure*}[t]
    \begin{tcolorbox}[
        enhanced,
        colback=white,
        colframe=black,
        arc=3mm,
        boxrule=0.5pt,
        title=Human Annotation Guideline.
    ]
    Annotators are asked to write a concise explanation of the video's implicit meaning, rather than a literal description of visible or audible content. The annotation should capture the core implicit meaning, emotion, satire, cultural reference, or rhetorical point conveyed by the video.

    \vspace{1ex}
    \textbf{Language rule.} Use the video's primary language, determined by speech first and then by visible text. If multiple languages appear, use the language most relevant to the implicit meaning. Do not translate into another language or mix languages unless the video itself relies on mixed-language cues.

    \vspace{1ex}
    \textbf{Content rule.} Focus on explanation rather than transcription. Avoid line-by-line paraphrase, unnecessary details, unsupported assumptions, or broad external analysis. Mention visual, textual, audio, timing, editing, or cultural cues only when they are necessary for explaining the underlying meaning.

    \vspace{1ex}
    \textbf{Style rule.} Write directly and briefly, preferably in 1 to 2 sentences and at most 3 sentences. Do not add titles, quotation marks, bullet points, prefaces, or meta-comments such as ``This video means that''.

    \vspace{1ex}
    \textbf{Ambiguity and sensitivity.} If the meaning is unclear, purely literal, random, or dependent on inaccessible context, flag the example for review. Sensitive content should be described neutrally and analytically, without endorsement or unnecessary repetition of offensive language.

    \end{tcolorbox}
    \caption{Human annotation guideline for implicit-meaning explanations.}
    \label{fig:annotation_guideline}
\end{figure*}

\subsection{Modality Annotation}
\label{app:modality-annotation}

Each clip carries a label naming the modalities its interpretation depends on. The three sources are on-screen text or captions (\textmod), audio including speech and non-speech sound (\audiomod), and visual content (\videomod), giving seven non-empty combinations; the tables write a combination as the symbols joined, so \textmod+\videomod\ is a clip whose point needs both the on-screen text and the picture. The label is not an inventory of what the clip contains. It records what the drivelological point rests on, which annotators decided by asking whether the point would still be recoverable with a stream removed: background music that could be muted without changing the reading does not make a clip audio-dependent, whereas a sound effect or a tone of voice that reverses the meaning of an otherwise ordinary visual event does. A clip can therefore be labelled as carried by on-screen text even though someone is speaking throughout it, and this is the usual case for the text-carried group.

The labels are distributed unevenly, because the phenomenon is: audio $650$ clips, audio and vision $183$, on-screen text $109$, text and vision $37$, text with audio and vision $11$, text and audio $6$, and vision alone $4$. Speech carries the drivelological point in most of the corpus, and the combinations that need all three streams at once are rare.

\subsection{Language Annotation}
\label{app:language-annotation}

We recorded language labels according to the linguistic signal used to infer the underlying meaning of each video. When speech was present, the speech-language label was determined by the spoken language or languages. When on-screen text contributed to interpretation, the caption-language label was determined by the language or languages of the relevant text, regardless of whether speech was also present. If neither speech nor on-screen text was present, the corresponding language field was marked as \textit{None}. When multiple languages contributed to interpretation, all relevant languages were recorded. 
The language labels are intended primarily for transparency. The dataset is not designed as a balanced multilingual benchmark, and performance comparisons across languages should therefore be interpreted cautiously.

\subsection{Ethical Considerations for Annotation}
\label{app:annotation-ethics}

Annotators were informed that the dataset may contain offensive, discriminatory, sexual, or otherwise sensitive content. Annotators could skip or flag examples that were difficult, ambiguous, or uncomfortable to annotate. Sensitive examples were retained only when they were relevant to the benchmark and were documented with appropriate content remarks. 
The dataset is intended for academic analysis of multimodal implicit meaning understanding. It should not be used to endorse, amplify, or generate harmful content. We include content warnings and sensitive-content metadata to support responsible use.

\section{Evaluation Details}
\label{app:evaluation-details}

\subsection{Explanation Prompt}
\label{app:generation-prompt}

For the explanation task on DrivelHub+, each model is given a video and prompted to produce a concise interpretation of its implicit Drivelological meaning. The prompt is designed to discourage surface-only captioning and instead elicit the video's underlying pragmatic point, such as its humour, satire, stance, affective implication, cultural reference, or rhetorical reversal. 
The prompt used in our experiments is shown in Figure~\ref{fig:generation_prompt}.

\begin{figure*}[t]
    \begin{tcolorbox}[
        enhanced,
        colback=white,
        colframe=black,
        arc=3mm,
        boxrule=0.5pt,
        title=Generation Prompt.
    ]
    Output only a short explanation of the video's core meaning or implied message. Do not add titles, quotation marks, introductions, bullet points, numbering, or extra formatting.

    \vspace{1ex}
    \textbf{Language rule.} Detect the video's primary language from spoken audio and on-screen text. The response must be written entirely in that same language. Do not translate, mix languages, or add explanations in another language.

    \vspace{1ex}
    \textbf{Content rule.} Explain the underlying meaning, emotion, joke, sarcasm, irony, or social implication of the video. Do not simply describe or summarise what happens scene by scene. Avoid repeating the video's exact words unless necessary, and focus on interpretation rather than transcription.

    \vspace{1ex}
    \textbf{Style rule.} Keep the response concise, usually 1 to 2 sentences and at most 3 short sentences. Write naturally as a single paragraph only. Do not use bullet points, lists, headers, labels, markdown, opening phrases such as ``This video shows...'' or ``The video means...'', or unnecessary analysis, background context, or moral commentary.

    \vspace{1ex}
    If the output language does not match the video's primary language, the response is considered incorrect.
    \end{tcolorbox}
    \caption{Prompt used for explanation evaluation.}
    \label{fig:generation_prompt}
\end{figure*}

\subsection{Video LLM-as-a-Judge Protocol}
\label{app:judge-protocol}

For explanation evaluation, we use a Video LLM-as-a-judge protocol to assess whether model-generated explanations capture the implicit Drivelological meaning of each video. Our judge design is inspired by the structured open-ended evaluation protocol used in ViMU \cite{li2026vimubenchmarkingvideometaphorical}, but is adapted to Drivelology interpretation by explicitly scoring whether a response captures the implicit narrative, the relevant rhetorical mechanism, the affective or social implication, and the multimodal evidence supporting the interpretation.

Given an input video, the judge receives two textual fields: the human-written annotation describing the video's implicit meaning, and the model-generated explanation. The judge is instructed to evaluate semantic understanding rather than lexical overlap. A response is considered aligned if it substantially preserves the same implicit narrative, emotional or social implication, and rhetorical interpretation as the human annotation, even when expressed with different wording.

The scoring ranges are intentionally asymmetric because the dimensions differ in their centrality to the task. Core Intent is assigned the largest range, $[0,5]$, because recovering the main implicit meaning is the primary objective of the benchmark. Rhetorical Signal is assigned an intermediate range, $[0,3]$, because identifying the mechanism that makes the video meaningful is important but secondary to recovering the implicit interpretation. Affective or Social Meaning and Grounding are assigned smaller ranges, $[0,2]$, because they capture supporting aspects of the explanation: whether the response preserves the relevant social or emotional implication and whether it uses appropriate video evidence. The penalty dimensions use analogous ranges: Hallucination and Literal-only errors can strongly invalidate an answer and are therefore scored on $[0,3]$, while vagueness is scored on $[0,2]$ because it usually weakens rather than fully reverses an otherwise plausible response.

The judge assigns scores along four positive dimensions:
\begin{itemize}
    \item \textbf{Core Intent} $[0,5]$: whether the response captures the main implicit meaning.
    \item \textbf{Rhetorical Signal} $[0,3]$: whether the response identifies the relevant Drivelological mechanism, such as misdirection, inversion, paradox, switchbait, irony, satire, metaphor, wordplay, contrast, or related rhetorical structure.
    \item \textbf{Affective or Social Meaning} $[0,2]$: whether the response captures the relevant emotional implication, social implication, cultural meaning, target, institution, group, or value judgment when supported by the video.
    \item \textbf{Grounding} $[0,2]$: whether the response correctly uses the relevant visual, textual, speech, audio, timing, editing, or cross-modal evidence.
\end{itemize}

The judge also assigns three penalty scores:
\begin{itemize}
    \item \textbf{Hallucination Penalty} $[0,3]$: penalises invented claims not grounded in the video, annotation, or available evidence.
    \item \textbf{Literal-only Penalty} $[0,3]$: penalises responses that remain at surface-level description and miss the implicit meaning.
    \item \textbf{Vague or Overgeneralisation Penalty} $[0,2]$: penalises responses that are too generic or non-committal to demonstrate understanding of the specific video.
\end{itemize}

To make the rubric operational, we use the score anchors in Table~\ref{tab:judge-score-anchors}. These anchors define how partial credit is assigned within each dimension. For example, a Core Intent score of 2 indicates that the response recognises only a broad topic or partially related implication, while a score of 3 indicates that it captures the general intended direction but misses an important twist, target, or pragmatic relation.

\begin{table*}[t]
\centering
\footnotesize
\setlength{\tabcolsep}{5pt}
\renewcommand{\arraystretch}{1.08}
\begin{tabular}{p{0.10\linewidth}p{0.84\linewidth}}
\toprule
\textbf{Score} & \textbf{Guideline} \\
\midrule

\multicolumn{2}{l}{\textbf{Core Intent}} \\
0 & Misses the implicit meaning; irrelevant, empty, or contradictory. \\
1 & Mentions only a broad topic without recovering the intended point. \\
2 & Partly related, but captures only a weak or generic implication. \\
3 & Captures the general direction but misses a key twist, target, or relation. \\
4 & Mostly captures the implicit meaning, with minor omissions or imprecision. \\
5 & Accurately captures the central implicit narrative or pragmatic point. \\
\midrule

\multicolumn{2}{l}{\textbf{Rhetorical Signal}} \\
0 & Does not identify the mechanism or gives an incorrect one. \\
1 & Mentions a broad mechanism but does not explain how it operates. \\
2 & Identifies the main mechanism but only partly connects it to evidence. \\
3 & Correctly explains the mechanism, e.g., reversal, misdirection, or incongruity. \\
\midrule

\multicolumn{2}{l}{\textbf{Affective or Social Meaning}} \\
0 & Misses or distorts the relevant emotional, social, or evaluative implication. \\
1 & Partly captures the implication, but the target, stance, or value is vague. \\
2 & Correctly captures the relevant affective or social implication. \\
\midrule

\multicolumn{2}{l}{\textbf{Grounding}} \\
0 & Provides no grounded evidence or relies on unsupported claims. \\
1 & Uses some evidence but misses the decisive modality, cue, or timing. \\
2 & Grounds the interpretation in the relevant visual, audio, textual, or timing cues. \\
\midrule

\multicolumn{2}{l}{\textbf{Hallucination Penalty}} \\
0 & No unsupported claims that affect the interpretation. \\
1 & Minor unsupported detail that does not substantially change the interpretation. \\
2 & Noticeable unsupported claim that affects part of the interpretation. \\
3 & Major hallucination that drives or contradicts the interpretation. \\
\midrule

\multicolumn{2}{l}{\textbf{Literal-only Penalty}} \\
0 & Goes beyond surface description and explains the implicit meaning. \\
1 & Contains some interpretation but remains partly descriptive. \\
2 & Mostly describes surface content, with only a weak implied meaning. \\
3 & Purely literal or transcript-like, with no implicit interpretation. \\
\midrule

\multicolumn{2}{l}{\textbf{Vague or Overgeneralised Penalty}} \\
0 & Specific to the video and its implicit meaning. \\
1 & Somewhat generic, but still contains video-specific interpretation. \\
2 & Too generic or non-committal to show understanding of the example. \\

\bottomrule
\end{tabular}
\caption{Score anchors for the Video LLM-as-a-judge rubric. Positive dimensions assign partial credit for recovering the implicit meaning, rhetorical mechanism, social or affective implication, and grounding evidence. Penalty dimensions subtract credit for hallucination, literal-only description, and vagueness.}
\label{tab:judge-score-anchors}
\end{table*}

The final judge score sums the four positive dimensions and subtracts the three penalty dimensions, with a maximum possible score of 12. In addition to this scalar score, the judge returns a binary alignment label indicating whether the generated explanation substantially captures the same implicit meaning as the human annotation. Empty, irrelevant, purely surface-level, or meaning-substituting responses are marked as unaligned. The full judge prompt is shown in Figure~\ref{fig:judge_prompt}.

The judge must be able to see and hear the clip it grades: many of these videos turn on a spoken pun or on a sound that contradicts the picture, and a judge without that evidence would be scoring an explanation against material it cannot check. Both of our judges therefore accept video with its soundtrack, Gemini 3.8 Flash for the main results and Qwen3-Omni-30B-A3B-Instruct for the open second grading. The judge receives the input video, the human-written annotation, and the model-generated explanation, and returns both a binary alignment label and rubric-based scores. Each judge grades every system, so the scoring criterion is held constant across the systems being compared.

\begin{figure*}[t]
    \begin{tcolorbox}[
        enhanced,
        colback=white,
        colframe=black,
        arc=3mm,
        boxrule=0.5pt,
        title=Prompt Template (Video LLM-as-a-Judge).
    ]
    You are grading answers for implicit multimodal understanding. Judge semantic understanding, not style.

    \vspace{0.5ex}
    You will be shown a video, a human \textbf{Annotation}, and a model-generated \textbf{Explanation}. Determine whether the Explanation captures the same implicit narrative, emotional/social implication, and rhetorical interpretation as the Annotation.

    \vspace{0.5ex}
    Do not require exact wording match. Mark the answer as unaligned if it is empty, irrelevant, surface-level, reverses the implication, substitutes another meaning, or introduces unsupported hallucinations.

    \vspace{0.5ex}
    Score the answer using the following dimensions:
    \begin{quote}
    \texttt{core\_intent}: [0, 5]\\
    \texttt{rhetorical\_signal}: [0, 3]\\
    \texttt{affective\_or\_social\_meaning}: [0, 2]\\
    \texttt{grounding}: [0, 2]\\
    \texttt{hallucination\_penalty}: [0, 3]\\
    \texttt{literal\_only\_penalty}: [0, 3]\\
    \texttt{vague\_or\_overgeneralised\_penalty}: [0, 2]
    \end{quote}

    Compute the total score as the sum of positive dimensions minus all penalties. The maximum possible score is 12.

    \vspace{0.5ex}
    Return only valid JSON with the following fields:
    \vspace{-0.5ex}
    \begin{verbatim}
{
  "aligned": true or false,
  "core_intent": int,
  "rhetorical_signal": int,
  "affective_or_social_meaning": int,
  "grounding": int,
  "hallucination_penalty": int,
  "literal_only_penalty": int,
  "vague_or_overgeneralised_penalty": int,
  "score_total": int,
  "reasoning_short": "one concise English sentence"
}
    \end{verbatim}
    \vspace{-1ex}

    \textbf{Annotation}: \{annotation\}

    \textbf{Explanation}: \{explanation\}

    Return only valid JSON.

    \end{tcolorbox}
    \caption{Prompt used for Video LLM-as-a-judge evaluation.}
    \label{fig:judge_prompt}
\end{figure*}

\subsection{Judge Robustness and an Open Substitute}
\label{app:judge-robustness}

The main results are graded by Gemini 3.8 Flash, a hosted model that belongs to none of the evaluated families. That independence costs reproducibility: a reader cannot rerun a paid API for free. We therefore graded the benchmark a second time with Qwen3-Omni-30B-A3B-Instruct, an open audio-capable model that runs locally, and report how far the open judge reproduces the proprietary grading. A judge scores one explanation per clip for every system in Table~\ref{tab:generation}, so a complete grading covers the 1,000 clips once for each of the eight evaluated models and once for the human baseline: 9,000 explanations. Both judges received the same videos, the same soundtrack, the same prompt and the same rubric, giving 9,000 paired judgements. Table~\ref{tab:judge-agreement} gives both gradings.

At the system level the two gradings agree: Pearson $r = 0.80$, Spearman $\rho = 0.70$, Kendall $\tau_b = 0.61$ over the nine systems. The top three positions are identical under both judges (i.e., the human baseline, then Qwen3-Omni-30B-A3B in thinking mode, then the same model without thinking) as is the direction of every comparison we draw in \S\ref{sec:results-discussion}. For individual explanations agreement is moderate but the strongest we measured for any pair of judges: $r = 0.53$ on the total and Cohen's $\kappa = 0.44$ on the binary aligned label. A reader without API access can therefore reproduce the benchmark's conclusions with the open judge, while scores for individual explanations should be treated as indicative rather than exact.

One system accounts for essentially all of the rank disagreement. MiniCPM-o-2.6 is placed $4$th of nine by the open judge and $9$th by the proprietary judge, a gap of $8.16$ against $0.74$ out of $12$; no other system moves more than two places. It is also the one system whose answers are far longer than the norm, at $321$ words against $24$ to $70$ for every other system. The mechanism is measurable rather than inferred: the open judge's totals rise with answer length (Spearman $\rho = +0.14$), the proprietary judge's fall ($\rho = -0.16$), and the difference between the two judges tracks length at $\rho = +0.25$. The open judge is therefore a safe substitute for systems that answer at ordinary length, and unreliable for systems that answer at several times the norm, where its length preference inflates them. We state the condition rather than the correlation alone, because it is the condition a reader needs in order to know when the substitution holds.

The proprietary judge is the stricter of the two. Gemini marks $36.8\%$ of responses aligned against $62.0\%$ for the open judge, and its totals are on average $4.13$ points lower. The strictness is not uniform flattery in reverse: it widens the gap the benchmark exists to measure, from $+1.67$ points between the human baseline and the best model under the open judge to $+2.64$ under the proprietary judge. A more lenient judge would have reported the task as closer to solved. The proprietary judge also scores the human baseline at $8.00$ rather than near $12$, chiefly on social meaning ($1.01$ of $2$) and grounding ($1.06$ of $2$), so the totals in Table~\ref{tab:generation} should be read as positions on a strict scale, not as fractions of attainable understanding.

Per-dimension agreement is highest on the dimension carrying most of the score, core intent ($r = 0.52$), followed by rhetorical signal ($r = 0.44$) and the literal-only penalty ($r = 0.40$). It is lowest on grounding ($r = 0.09$): the judges concur on whether a response recovered the intended meaning far more than on whether it used the right evidence to get there. Deciding what counts as correct multimodal grounding is evidently the least transferable judgement in the rubric, and conclusions that depend on that dimension alone should be treated with more caution than the totals.

Table~\ref{tab:judge-cost} reports the cost of every grading in this paper, priced from the recorded token counts at the published rates: \$$24.21$ for one complete pass over the benchmark, \$$7.95$ for the cascaded baseline, and \$$32.16$ for all $12{,}000$ judgements, or \$$0.0027$ each. The judge receives every clip in full with its soundtrack, so input tokens scale with clip duration and are almost independent of the response being graded: seven of the nine gradings cost between \$$2.64$ and \$$2.66$, within $1\%$ of each other. The exceptions are Qwen2.5-Omni-3B at \$$2.72$ and MiniCPM-o-2.6 at \$$2.96$, whose $321$-word answers add roughly $400$ input tokens each (i.e., the same verbosity that separates the two judges also makes it the most expensive system to evaluate, by a small margin). Because the video dominates, the cost of applying this benchmark scales with the corpus, not with the systems under test: each further system costs another \$$2.65$, and a longer video corpus costs proportionally more. These are published-rate costs, which is what another group can predict before committing to a run. Our own billing came to about a quarter more, GBP $31.77$ for the $12{,}000$ judgements in this paper against a listed \$$32.16$, a gap that is usual for video input; we report the listed figure because the excess carries an exchange rate and a tax jurisdiction that do not transfer to anyone else. The open judge of this appendix has no monetary cost at all, which is the other half of the case for reporting it.

\begin{table}[t]
\centering
\small
\resizebox{\linewidth}{!}{
\begin{tabular}{lrrr}
\toprule
\textbf{Graded system} & \textbf{In} & \textbf{Out} & \textbf{USD} \\
\midrule
MiniCPM-o-2.6 9B & 3,244 & 139 & 2.96 \\
Qwen2.5-Omni 3B & 2,928 & 140 & 2.72 \\
Qwen3-Omni 30B-A3B (No-thinking) & 2,845 & 140 & 2.66 \\
Qwen2.5-Omni 7B & 2,834 & 140 & 2.65 \\
EchoInk-R1 7B & 2,833 & 140 & 2.65 \\
Qwen3-Omni 30B-A3B (Thinking) & 2,834 & 140 & 2.65 \\
Gemma 4 E4B & 2,821 & 140 & 2.64 \\
Human baseline & 2,833 & 137 & 2.64 \\
Gemma 4 E2B & 2,819 & 140 & 2.64 \\
\midrule
\textit{One full pass, 9,000 judgements} & & & \textit{24.21} \\
\midrule
Cascade: OCR only & 2,836 & 141 & 2.65 \\
Cascade: OCR + ASR & 2,834 & 140 & 2.65 \\
Cascade: ASR only & 2,829 & 140 & 2.65 \\
\midrule
\textit{Cascaded baseline, 3,000 judgements} & & & \textit{7.95} \\
\midrule
\textbf{All 12,000 judgements} & & & \textbf{32.16} \\
\bottomrule
\end{tabular}
}
\caption{Cost of every Gemini 3.8 Flash grading in this paper, at the published token prices: the 9,000 explanations of Table~\ref{tab:generation}, and the 3,000 cascade responses of Table~\ref{tab:cascade}. \textbf{In} and \textbf{Out} are mean input and output tokens per judgement; input is dominated by the video, which the judge receives in full with its soundtrack. At \$0.0027 per judgement, re-grading the benchmark costs about \$24. The open judge of Appendix~\ref{app:judge-robustness} costs nothing but GPU time. Billing for this project ran about a quarter above the listed rate, which is the usual gap for video input.}
\label{tab:judge-cost}
\end{table}

Table~\ref{tab:generation-qwen3-omni-30b-a3b-instruct} gives the open judge's full grading, with the same rubric breakdown as Table~\ref{tab:generation}. Reading the two side by side shows where the difference between the judges sits: the open judge is more generous on every positive dimension, most of all on grounding ($1.76$ against $1.01$ of $2$), and assigns smaller penalties, so its totals are higher throughout without the ordering changing much. A reader who wants to reproduce the benchmark locally can compare against this table directly.

\begin{table*}[t]
\centering
\resizebox{\linewidth}{!}{
\begin{tabular}{lllccccccccc}
\toprule
\textbf{Model} 
& \textbf{Size}
& \textbf{Mode}
& \textbf{Aligned\%} $\uparrow$
& \textbf{Core/5} $\uparrow$
& \textbf{Rhet./3} $\uparrow$
& \textbf{Social/2} $\uparrow$
& \textbf{Ground./2} $\uparrow$
& \textbf{Halluc./3} $\downarrow$
& \textbf{Literal/3} $\downarrow$
& \textbf{Vague/2} $\downarrow$
& \textbf{Total/12} $\uparrow$ \\
\midrule
\textcolor{gray}{Human baseline} & \textcolor{gray}{--} & \textcolor{gray}{--}
& \textcolor{gray}{0.993} & \textcolor{gray}{4.976} & \textcolor{gray}{2.988} & \textcolor{gray}{1.942} & \textcolor{gray}{1.996} & \textcolor{gray}{0.008} & \textcolor{gray}{0.006} & \textcolor{gray}{0.000} & \textcolor{gray}{11.888} \\
\midrule

Qwen3-Omni & 30B-A3B & Thinking
& \bestA{0.805} & \bestA{4.394} & \bestA{2.713} & \bestB{1.708} & \bestB{1.915} & \bestA{0.294} & \bestA{0.176} & \bestA{0.044} & \bestA{10.216} \\
Qwen3-Omni & 30B-A3B & No-thinking
& \bestB{0.798} & \bestB{4.377} & \bestB{2.697} & \bestA{1.710} & \bestA{1.921} & \bestC{0.344} & \bestB{0.186} & \bestB{0.049} & \bestB{10.126} \\
MiniCPM-o-2.6 & 9B & Instruct
& \bestC{0.713} & \bestC{3.722} & \bestC{2.260} & \bestC{1.469} & \bestC{1.816} & \bestB{0.301} & \bestC{0.750} & \bestC{0.054} & \bestC{8.162} \\
EchoInk-R1 & 7B & Instruct
& 0.533 & 3.282 & 2.093 & 1.225 & 1.754 & 0.492 & 0.917 & 0.204 & 6.741 \\
Qwen2.5-Omni & 7B & Instruct
& \worstC{0.502} & 3.176 & 2.051 & \worstC{1.175} & 1.726 & \worstC{0.571} & 0.918 & \worstC{0.206} & 6.433 \\
Qwen2.5-Omni & 3B & Instruct
& 0.528 & \worstC{3.118} & \worstC{1.984} & 1.185 & \worstC{1.692} & 0.549 & \worstB{1.049} & 0.188 & \worstC{6.193} \\
Gemma 4 & E4B & No-thinking
& \worstB{0.417} & \worstB{2.728} & \worstB{1.778} & \worstB{1.017} & \worstB{1.574} & \worstA{0.751} & \worstC{1.035} & \worstB{0.284} & \worstB{5.027} \\
Gemma 4 & E2B & No-thinking
& \worstA{0.293} & \worstA{2.129} & \worstA{1.462} & \worstA{0.753} & \worstA{1.464} & \worstB{0.697} & \worstA{1.419} & \worstA{0.442} & \worstA{3.250} \\

\bottomrule
\end{tabular}
}
\caption{
Video LLM-as-a-judge results for implicit-meaning understanding on 1,000 drivelological videos, graded by Qwen3-Omni-30B-A3B-Instruct. Green shading marks the top three results in each column and purple shading the bottom three, computed over the models only; the human baseline is the reference ceiling and is excluded from the shading.
}
\label{tab:generation-qwen3-omni-30b-a3b-instruct}
\end{table*}

\begin{table*}[t]
\centering
\small
\begin{tabular}{llcccccc}
\toprule
\multirow{2}{*}{\textbf{Model}} & \multirow{2}{*}{\textbf{Setting}}
& \multicolumn{3}{c}{\textbf{Gemini 3.8 Flash}}
& \multicolumn{3}{c}{\textbf{Qwen3-Omni-30B-A3B}} \\
\cmidrule(lr){3-5} \cmidrule(lr){6-8}
& & Total/12 & Aligned\% & Rank & Total/12 & Aligned\% & Rank \\
\midrule
Human baseline & -- & 8.00 & 99.7 & 1 & 11.89 & 99.3 & 1 \\
\midrule
Qwen3-Omni 30B-A3B & Thinking & 5.36 & 61.4 & 2 & 10.22 & 80.5 & 2 \\
Qwen3-Omni 30B-A3B & No-thinking & 4.98 & 56.5 & 3 & 10.13 & 79.8 & 3 \\
EchoInk-R1 7B & Instruct & 3.11 & 32.8 & 4 & 6.74 & 53.3 & 5 \\
Qwen2.5-Omni 7B & Instruct & 3.02 & 30.4 & 5 & 6.43 & 50.2 & 6 \\
Gemma 4 E4B & No-thinking & 2.31 & 18.7 & 6 & 5.03 & 41.7 & 8 \\
Gemma 4 E2B & No-thinking & 1.69 & 10.2 & 7 & 3.25 & 29.3 & 9 \\
Qwen2.5-Omni 3B & Instruct & 1.64 & 15.6 & 8 & 6.19 & 52.8 & 7 \\
MiniCPM-o-2.6 9B & Instruct & 0.74 & 6.1 & 9 & 8.16 & 71.3 & 4 \\
\bottomrule
\end{tabular}
\caption{The same systems graded by Gemini 3.8 Flash and Qwen3-Omni-30B-A3B. Each judge scores one explanation per clip for every system in the table, 9,000 judgements per judge. The first column pair is the proprietary grading used in Table~\ref{tab:generation}; the human baseline is the reference ceiling.}
\label{tab:judge-agreement}
\end{table*}

\begin{figure}[t]
\centering
\includegraphics[width=\columnwidth]{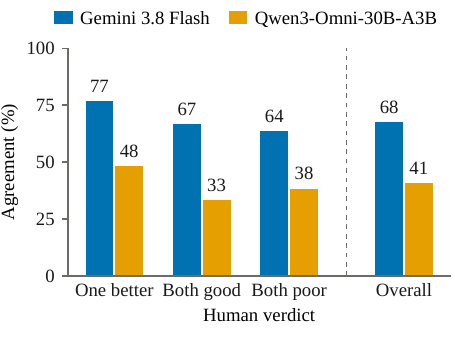}
\caption{Agreement between each judge and a human annotator, grouped by the human verdict on a pair of explanations. A judge's verdict comes from its aligned label for each explanation. It finds one better when it marks only one aligned, both good when it marks both, and both poor when it marks neither.}
\label{fig:human-alignment}
\end{figure}

To test whether the judges agree with people, two human annotators compared $200$ pairs of explanations of the same clip and judged whether one was better, both were good, or both were poor. Each judge gives the same verdict through its aligned label for the two explanations. Figure~\ref{fig:human-alignment} shows that Gemini 3.8 Flash matches the human far more often than Qwen3-Omni-30B-A3B, at $68\%$ against $41\%$. Cohen's $\kappa$, which measures agreement beyond what chance alone would produce, is $0.41$ for Gemini and $0.13$ for Qwen3-Omni. The judges rarely dispute which explanation is better, reversing a human preference on only $3$ pairs between them. Where they differ from the human is on whether an explanation is good enough. Qwen3-Omni accepts many explanations the annotator rejected, which is the source of its leniency reported above. The annotator also judged both explanations poor in $138$ of the $200$ pairs, which confirms that current systems often miss the implicit meaning. Gemini follows human judgement most closely, which supports using it for the main results.

\subsection{A Cascaded Text-Conversion Baseline}
\label{app:cascade}

A multimodal model that scores poorly here could be failing for three different reasons: it may not perceive the evidence, it may perceive it but fail to extract the relevant channel, or it may extract everything and still not infer the implicit meaning. Aggregate scores cannot separate these. We therefore convert each clip to text with off-the-shelf components and give that text, and nothing else, to a language model.

Speech is transcribed with Whisper large-v3 \cite{radford2023robust} and on-screen text is read with EasyOCR \cite{easyocr} over 12 uniformly sampled frames, with the recogniser's language set chosen per clip from the annotated caption language. The resulting text is given to Qwen3-Omni-30B-A3B-Instruct with no frames and no audio. Using the same model family that also appears among the evaluated systems is deliberate: it makes the comparison an ablation of the input representation rather than a comparison of reasoning ability, so the gap between the cascade and the multimodal run is attributable to what the pipeline discards. Three conditions isolate the channels: the transcript alone, the on-screen text alone, and both. Table~\ref{tab:cascade} reports all three against the multimodal model, graded on the same rubric by the same judge as Table~\ref{tab:generation}, with the open judge used as a check.

The columns of that table group the clips by the modality annotation of Appendix~\ref{app:modality-annotation}, that is, by the stream the drivelological point rests on rather than by the streams the clip contains. The grouping is what makes the baseline diagnostic rather than merely another score: a condition that supplies the carrying stream should track the multimodal model, and the same condition should fail on clips carried by a stream it discards, so the two cases can be told apart. We report the four labels holding at least $15$ clips, namely audio ($650$ clips), audio and vision ($183$), on-screen text ($109$), and text and vision ($37$). The remaining $21$ clips fall into three combinations too small to read anything from and so get no column of their own; they are still counted in the All column, which is why the four groups shown do not sum to 1,000.

The full cascade reaches $4.00$ of $12$ against the multimodal model's $4.98$, or $80\%$. On the $650$ clips whose meaning is carried by speech it reaches $4.64$ against $5.40$, or $86\%$. A text pipeline therefore recovers most of what this benchmark measures on most of its clips, and for those clips the difficulty is not pragmatic inference that only a multimodal model can perform. It is available in the transcript, and a text model does nearly as well with it. 
Grading the same $3{,}000$ cascade responses with the open judge gives the same answer, and Table~\ref{tab:cascade} reports both. Its absolute scores are far higher, because it is the more lenient grader and puts the multimodal model at $10.13$ rather than $4.98$, but the proportion recovered is almost unchanged: $63\%$ for the transcript alone, $44\%$ for the on-screen text alone, and $82\%$ for both, against $63\%$, $44\%$ and $80\%$ here. What the cascade recovers is therefore a property of the pipeline rather than of the scale a particular judge happens to use. 

Each condition fails exactly where its modality is absent. The transcript alone scores $4.18$ on audio-carried clips and $-0.39$ on text-carried ones; the on-screen text alone reverses this, at $2.32$ and $3.63$. The negative total is not a rounding artefact: given a transcript that does not contain the joke, the model invents a reading, and the rubric's hallucination and literal-only penalties exceed everything it earns. The open judge shows the same reversal on a wider scale, at $9.23$ and $-4.87$ for the transcript and $5.03$ and $7.33$ for the on-screen text. These are failures of modality extraction, diagnosable as such because the condition that supplies the missing channel recovers the clip.

The residual is visible where vision participates. Adding vision to audio costs the multimodal model $5.40 \rightarrow 4.11$, a drop of $24\%$, and costs the full cascade $4.64 \rightarrow 2.78$, a drop of $40\%$. On the $37$ clips carrying meaning in both text and vision the cascade falls to $0.95$ against the multimodal model's $3.38$, keeping $28\%$ where it keeps $80\%$ overall. That gap is the genuinely multimodal part of the benchmark: meaning carried by a frame-level change, by timing, or by the relation between what is shown and what is said, which survives neither transcription nor optical character recognition. It is a minority of the corpus, and it is the part no text pipeline reaches.

Whisper hallucinates on silence: $54$ of the $138$ clips annotated as having no speech receive a transcript anyway, $43$ of them a canonical artefact such as \textit{Thank you for watching} and $11$ a degenerate repetition. We leave these in, because suppressing them using the speech annotation would leak the ground truth into the baseline and flatter the cascade; an off-the-shelf pipeline really does behave this way, and the effect is confined to clips whose meaning the transcript could not have carried. The OCR half uses a dedicated recogniser rather than a vision-language model prompted to transcribe, so these numbers are a lower bound on what a stronger text extractor would achieve. We did not run the video-captioning-then-LLM variant, which would test a fourth path from pixels to text.

\begin{table*}[t]
\centering
\small
\resizebox{\linewidth}{!}{
\begin{tabular}{lrrrrrrrrrr}
\toprule
\multirow{2}{*}{\textbf{Input to the model}} & \multicolumn{5}{c}{\textbf{Gemini 3.8 Flash}} & \multicolumn{5}{c}{\textbf{Qwen3-Omni-30B-A3B}} \\
\cmidrule(lr){2-6} \cmidrule(lr){7-11}
 & \textmod \audiomod \videomod & \audiomod & \audiomod \videomod & \textmod & \textmod \videomod & \textmod \audiomod \videomod & \audiomod & \audiomod \videomod & \textmod & \textmod \videomod \\
\midrule
Video with soundtrack (multimodal) & 4.98 & 5.40 & 4.11 & 4.71 & 3.38 & 10.13 & 10.79 & 9.02 & 9.09 & 7.43 \\
\midrule
ASR only & 3.15 & 4.18 & 2.60 & -0.39 & -0.76 & 6.33 & 9.23 & 5.51 & -4.87 & -4.46 \\
OCR only & 2.17 & 2.32 & 1.16 & 3.63 & 0.92 & 4.47 & 5.03 & 1.83 & 7.33 & 1.57 \\
OCR + ASR & 4.00 & 4.64 & 2.78 & 3.63 & 0.95 & 8.31 & 9.70 & 6.14 & 6.77 & 1.49 \\
\bottomrule
\end{tabular}
}
\caption{The cascaded text-conversion baseline under both judges, over the 1,000 clips graded in every setting. Columns after the first group the clips by the stream the drivelological point rests on, using the modality annotation of Appendix~\ref{app:modality-annotation}: \audiomod\ audio, \textmod\ on-screen text, \videomod\ vision, and a combination where the point needs both. The groups shown are those holding at least 15 clips, and their sizes are given in the text. The two judges disagree on the scale and agree on the finding: the open judge scores everything roughly twice as high, but a condition tracks the multimodal model where its own modality carries the meaning and collapses where it does not under both. Negative totals arise where the rubric's penalties exceed its positive dimensions, as when a model confabulates a reading from a transcript that does not contain the drivelology.}
\label{tab:cascade}
\end{table*}

\subsection{Where the Multimodal Systems Fail}
\label{app:vision-gap}

The cascade above shows a text pipeline collapsing wherever vision participates. On its own that could be a fact about the pipeline rather than about the clips, so we put the same question to the multimodal systems, which discard nothing. Figure~\ref{fig:vision-gap} splits each system's alignment rate by whether the modality annotation of Appendix~\ref{app:modality-annotation} says vision is needed to recover the point. Every one of the eight is less accurate on the clips that need vision, under both judges: by between $2$ and $18$ points under the proprietary judge, and between $14$ and $24$ under the open one. The direction is the same for every system and every judge, so it does not depend on which model grades the answers.

The human baseline is the control, and it shows no gap at all: $100.0\%$ against $99.6\%$ under the proprietary judge, $99.1\%$ against $99.3\%$ under the open one. The clips that need vision are therefore not intrinsically more ambiguous, nor more weakly annotated, nor harder to put into words; a person reading them writes an aligned explanation just as often. They are harder only for the models. Read together with the cascade, the two experiments converge from opposite directions on the same subset: what resists a text pipeline and what resists a multimodal system are both concentrated where the drivelological point is carried by what is shown rather than by what is said or written.

\begin{figure*}[t]
    \centering
    \includegraphics[width=\textwidth]{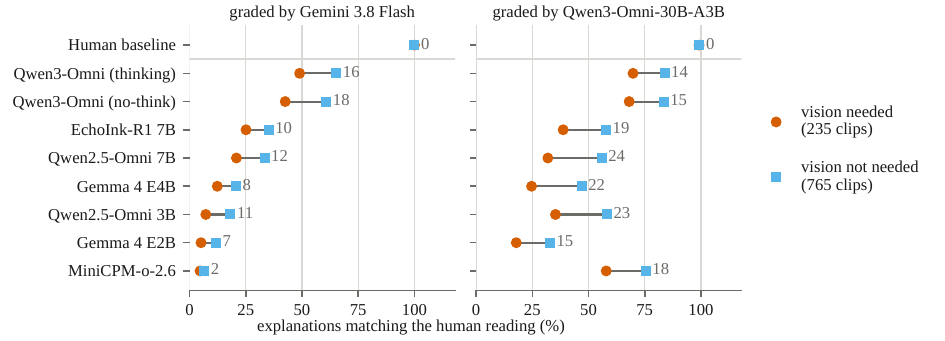}
    \caption{%
        Alignment rate of every evaluated system, split by whether the modality annotation says vision is needed to recover the implicit meaning. Each row is one system: the circle is its rate on the $235$ clips that need vision, the square its rate on the remaining $765$, and the number to the right is the gap in percentage points. Every system loses accuracy when vision carries the meaning, under the proprietary judge and the open one alike; the panels are named by the model that graded them, not by a graded system. The human baseline, above the rule, is the control and shows no gap, which rules out those clips simply being harder to interpret or more weakly annotated.
    }
    \label{fig:vision-gap}
\end{figure*}

\subsection{Writing-Style Measures}
\label{app:style}

Both measures in Section~\ref{sec:results-discussion} count word tokens and word $n$-grams, so both run over the $877$ clips whose human annotation is written in the Latin alphabet, and over the same clip set for every source. The other $123$ are annotated wholly or partly in Chinese, Japanese, Korean or Thai, where whitespace does not delimit words the same way and token counts would not be comparable across sources. The restriction is on the language of the annotation, not of the clip: the corpus carries speech in more than a dozen languages, and a clip with Mandarin speech is included whenever its annotation is written in English.

For reused phrasing, each source's reference is a random sample of its own explanations drawn until a fixed word budget is reached, and the score for an explanation outside that sample is the share of its word trigrams occurring in it. The budget matters: MiniCPM-o-2.6 writes about ten times as many words per explanation as a human annotator, so without a cap any phrase of its would have ten times the opportunity to recur, and the measure would rank verbosity rather than repetition. The figure averages $20$ such reference draws, which is also not optional. A verbose source fills the budget from only about $30$ documents, and single draws placed MiniCPM-o-2.6 anywhere between $10.4\%$ and $24.4\%$, a swing wider than most differences between models. The source ordering is insensitive to the $n$-gram order (Spearman $\rho \geq 0.945$ for $n = 2$ to $5$), but the separation is not: at $n = 2$ bigrams are common enough that every source reuses them and the human annotations stop being lowest, while by $n = 5$ most medians sit at zero. Only $n = 3$ and $n = 4$ place both human sets below every model, and $n = 4$ already floors two sources, so we report $n = 3$.

For the discriminant projection, each explanation is described by the relative frequencies of $150$ common function words, punctuation rates, and four structural ratios, all normalised by length so that the space is near-independent of topic and of how much a source chose to write. A single $33$-word explanation carries too little signal to place on its own, so the unit is a bundle of $25$ explanations from one source and its position is a sample mean. Clips are split in half, the projection is fitted on one half and only bundles drawn from the other are plotted. A logistic regression on the same features, cross-validated over five folds, separates human from model explanations at macro-$F_1$ $0.81$ and attributes a bundle to one of the ten sources well above chance.

\subsection{Retrieval Embedding Prompt}
\label{app:retrieval-prompt}

For representation-level retrieval on DrivelHub+, we follow the embedding extraction format used by LCO-style decoder-based embedding models \cite{xiao2026scaling}. Each input is wrapped with a short instruction prompt before computing its embedding. For text inputs, we use the following template:
\begin{quote}
\small
\texttt{\{text\}}\\
\texttt{Summarise the above text in one word:}
\end{quote}

For video inputs, the same instruction is paired with the video input in the official model inference format:
\begin{quote}
\small
\texttt{Summarise the above video in one word:}
\end{quote}

The embedding is then extracted from the model output representation following the official implementation or the model's recommended embedding interface. This prompt is not used to generate a natural-language answer; instead, it serves as an instruction for producing a compact representation suitable for similarity-based retrieval. We use the same prompt format for both text-to-video and video-to-text retrieval for models requiring instruction-formatted embeddings: LCO-Omni models and all the generative models.

\section{Case Study}

\subsection{Frame-Level Visual Evidence}
\label{app:reasoning-trace-case-study}

\begin{figure*}[tp]
    \centering
    \includegraphics[width=0.95\linewidth]{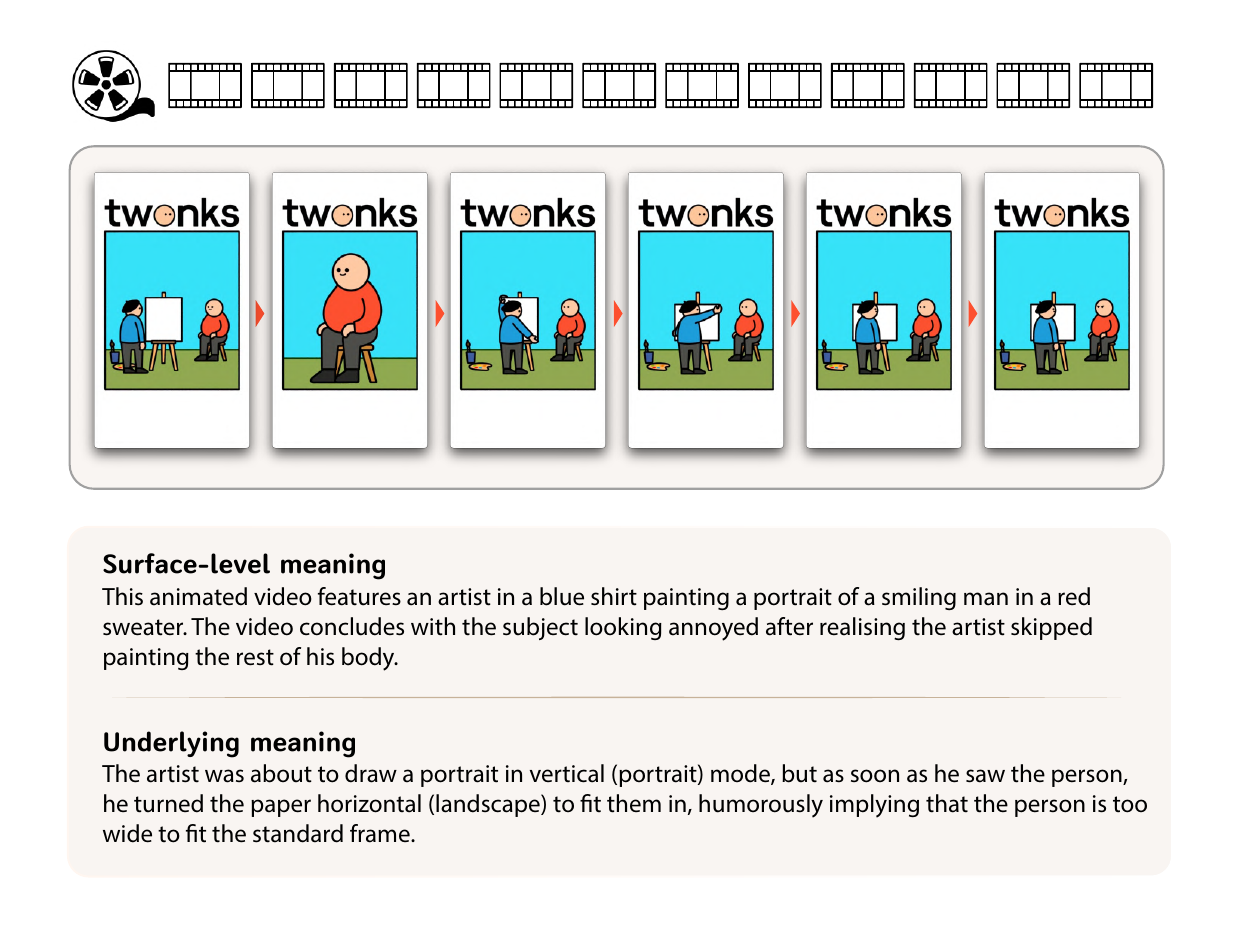}
    \caption{Example of a drivelological video requiring visual inference and frame-based humour.}
    \label{fig:example_vision}
\end{figure*}

The decisive evidence for a drivelological video is sometimes a single frame-level transition, and that transition is easy to lose. The example in Figure~\ref{fig:example_vision} is a visual pun: the artist starts with the paper in portrait orientation and turns it sideways the moment he sees the sitter, implying that the subject is too wide for a vertical frame.

None of the eight evaluated systems reports the rotation. MiniCPM-o-2.6 writes the longest answer any model produced for this clip and spends it on the colour of the shirts, the easel, the background, and the on-screen word ``twonks''; the canvas never changes orientation in its account, and the judge scores it $0$ with a literal-only penalty. Qwen3-Omni-30B-A3B in thinking mode, the strongest system in Table~\ref{tab:generation}, is concise and fluent, but substitutes a different story in which the sitter dislikes the finished artwork. The human annotator names the cue in one sentence and is the only aligned response on this clip, at $8$ out of $12$.

The failure is therefore neither verbosity nor brevity. Both models recognise the scene, an artist, a canvas and a seated person; what neither recovers is the one moment in which the paper turns, which is where the joke lives. Pragmatic interpretation of this example does not require more visual detail, but attention to the single transformation that carries the meaning.

\subsection{Score-Space Separation}
\label{app:score-space}

\begin{figure*}[t]
    \centering
    \includegraphics[width=\textwidth]{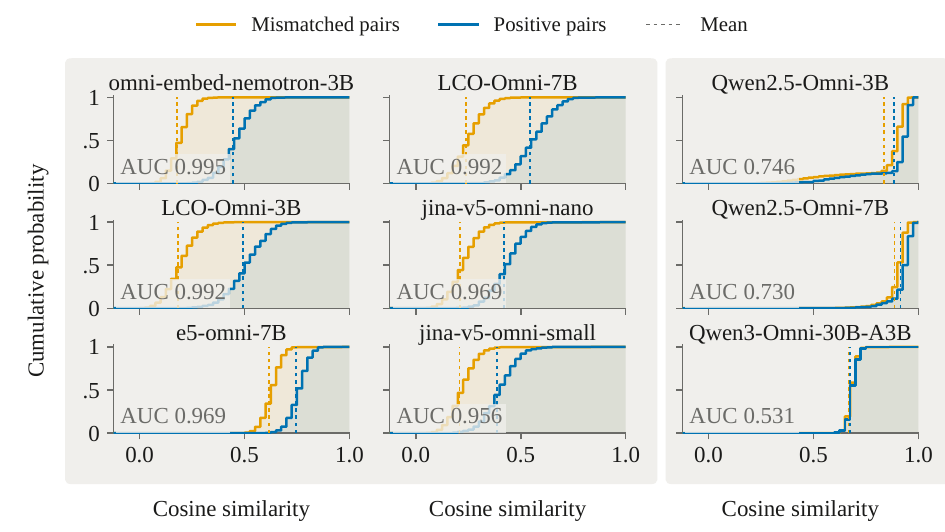}
    \caption{%
        Score distributions for positive and mismatched video-text pairs. Panels share common axes; dashed lines indicate means and models are ordered within each family by random-pair AUC. Pairwise separation is diagnostic only and should not be interpreted as R@1, which depends on query-level hardest negatives.
    }
    \label{fig:similarity-cdfs}
\end{figure*}

The retrieval metrics in Table~\ref{tab:retrieval} depend on relative rankings rather than on the absolute magnitude of cosine similarity. Figure~\ref{fig:similarity-cdfs} examines this distinction by comparing, within each model, the cumulative score distributions of positive and mismatched video-text pairs, where a positive pair is a clip and its own written interpretation. The embedding-native models show a clear rightward shift for positive pairs, whereas the generative models adapted for retrieval place positive and mismatched pairs in substantially overlapping score ranges. A high cosine value alone is therefore not evidence that a model can identify the correct match. The pairwise AUC shown in each panel measures how often a positive pair scores above a mismatched pair, but it weights all mismatches equally and is dominated by the bulk of easier negatives. R@1 is stricter: a query succeeds only when its positive item also outranks its highest-scoring mismatched candidate. We therefore use the CDFs to diagnose score-space separation and Table~\ref{tab:retrieval} to report retrieval performance.

\subsection{Direction-Specific Retrieval Failures}
\label{app:retrieval-direction-cases}

Table~\ref{tab:retrieval} places the two LCO-Omni checkpoints within about a point of each other, and the order changes with the direction: the 7B model leads text-to-video, $83.5$ against $82.2$ R@1, and the 3B model leads video-to-text, $80.5$ against $79.1$. That near-tie is not two models behaving alike. Table~\ref{tab:retrieval-direction-cases} gives the rank of the gold document for four clips under both checkpoints and both directions, and on every one of them at least one setting is more than a hundred places from the top. Ranks here are per-query positions of the best-matching gold document, which is a finer-grained quantity than the R@1 of Table~\ref{tab:retrieval}; the latter is a recall over all documents marked relevant for a query, so the two are not directly comparable.

\begin{table*}[t]
\centering
\small
\begin{tabular}{llcc}
\toprule
\multirow{2}{*}{\textbf{Case}}
& \multirow{2}{*}{\textbf{Direction}}
& \multicolumn{2}{c}{\textbf{Rank of the gold document}$\downarrow$} \\
\cmidrule(lr){3-4}
& & LCO-Omni-3B & LCO-Omni-7B \\
\midrule
\multirow{2}{*}{Diabetes \textit{type 1} read as a keystroke}
& Text-to-video & 320 & 168 \\
& Video-to-text & \textbf{3} & \textbf{1} \\
\midrule
\multirow{2}{*}{\textit{Watch} read as look, not mind}
& Text-to-video & 312 & 168 \\
& Video-to-text & 53 & \textbf{3} \\
\midrule
\multirow{2}{*}{Elaborate guessing system, first guess right}
& Text-to-video & 23 & \textbf{1} \\
& Video-to-text & 161 & 152 \\
\midrule
\multirow{2}{*}{Mother's advice dated by her own age}
& Text-to-video & \textbf{3} & 223 \\
& Video-to-text & \textbf{7} & 86 \\
\bottomrule
\end{tabular}
\caption{Per-query divergences between LCO-Omni-3B and LCO-Omni-7B, reported as the rank of the best-matching gold document among the 1,000 candidates. Lower is better, and ranks within the top three are set in bold. The first three cases separate the two retrieval directions: the same embeddings place the pair near the top one way and past rank 150 the other way. The fourth separates the two checkpoints instead, and it is the smaller one that succeeds.}
\label{tab:retrieval-direction-cases}
\end{table*}

On the first three cases it is the direction, not the checkpoint, that decides the outcome. A captioned exchange in which someone confesses a diagnosis of \textit{type 1} diabetes and the reply is to press the \textit{1} key is retrieved from the video at ranks $3$ and $1$, and from the written interpretation at ranks $320$ and $168$. A request to \textit{watch} someone's child, answered by looking at the child rather than minding it, behaves the same way, at $53$ and $3$ from the video against $312$ and $168$ from the text. The third case runs the other way: a friend explains an elaborate alphabetical guessing system and the first guess is accepted immediately, making the rules pointless, and here the written interpretation finds the clip at ranks $23$ and $1$ while the clip fails to find the interpretation at $161$ and $152$.

A reading consistent with these three, though four cases cannot establish it, is that the direction which succeeds is the one whose query side carries the distinctive material. The first two clips hold their wordplay in on-screen text that is unusual enough to locate, whereas their written interpretations are ordinary pragmatic prose, of the form \textit{a term is taken in its literal sense instead of its intended one}, which describes a great many clips in the corpus and singles out none of them. The third reverses this: the interpretation names an unusual object, an alphabetical hot-or-cold guessing system, whereas the clip is annotated as carrying its meaning in speech and captions, with nothing visually distinctive for the video encoder to fix on.

The fourth case separates the checkpoints rather than the directions. A $30$-year-old mother tells her $15$-year-old son to use protection and then realises she was $15$ when she had him; LCO-Omni-3B places the pair at ranks $3$ and $7$, and LCO-Omni-7B at $223$ and $86$. The aggregate near-tie therefore also conceals clips that one checkpoint handles and the other does not, and the checkpoint that succeeds is not the larger one.

This is why we report the two directions separately rather than averaging them. Text-to-video asks whether a written pragmatic interpretation can single out one clip among many similar ones; video-to-text asks whether the clip can be encoded close to that interpretation at all. A single averaged score would present these two checkpoints as near-equivalent and hide both failure modes.

\begin{figure*}[tp]
    \centering
    \includegraphics[width=0.95\linewidth]{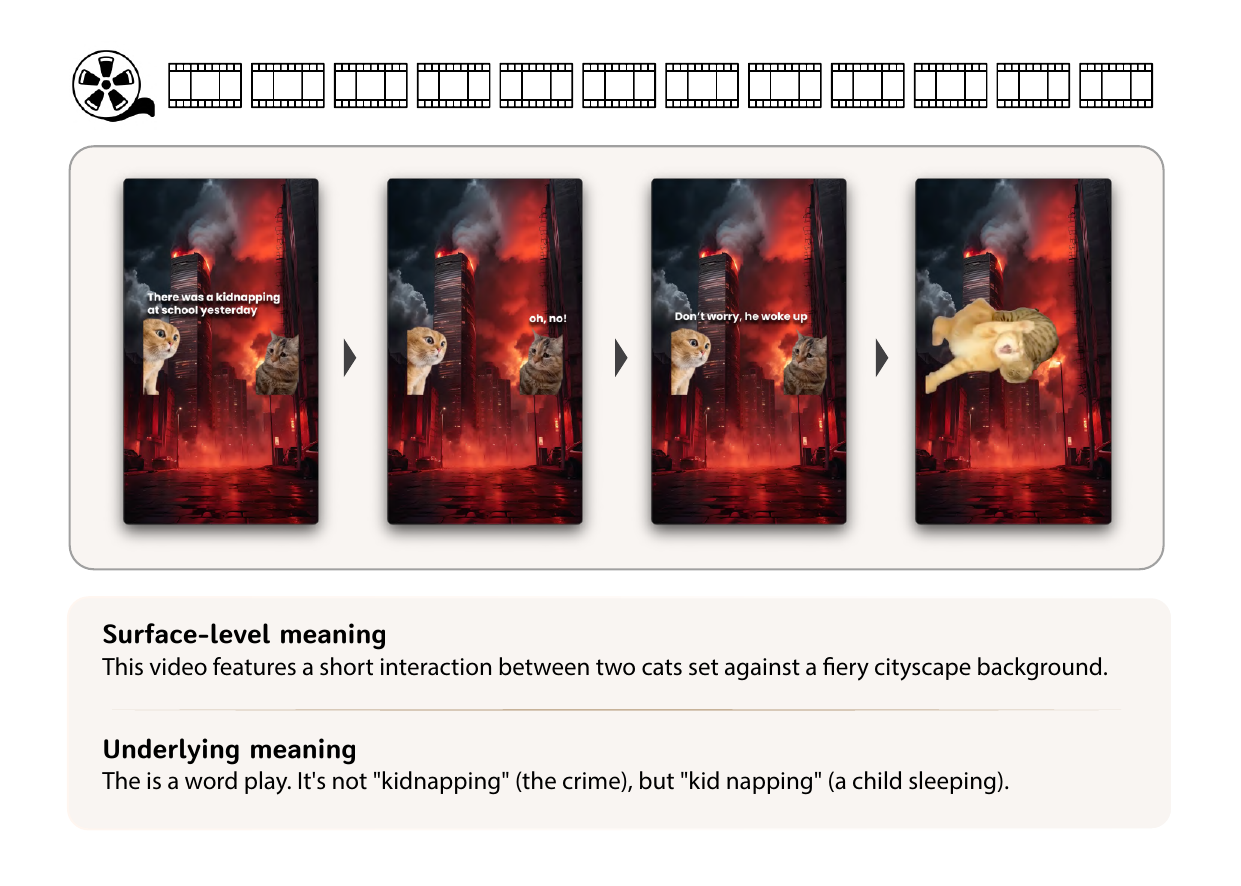}
    \caption{Example of a drivelological video that requires on-screen wordplay but does not require cross-modal interpretation.}
    \label{fig:example_text}
\end{figure*}

\begin{figure*}[tp]
    \centering
    \includegraphics[width=0.95\linewidth]{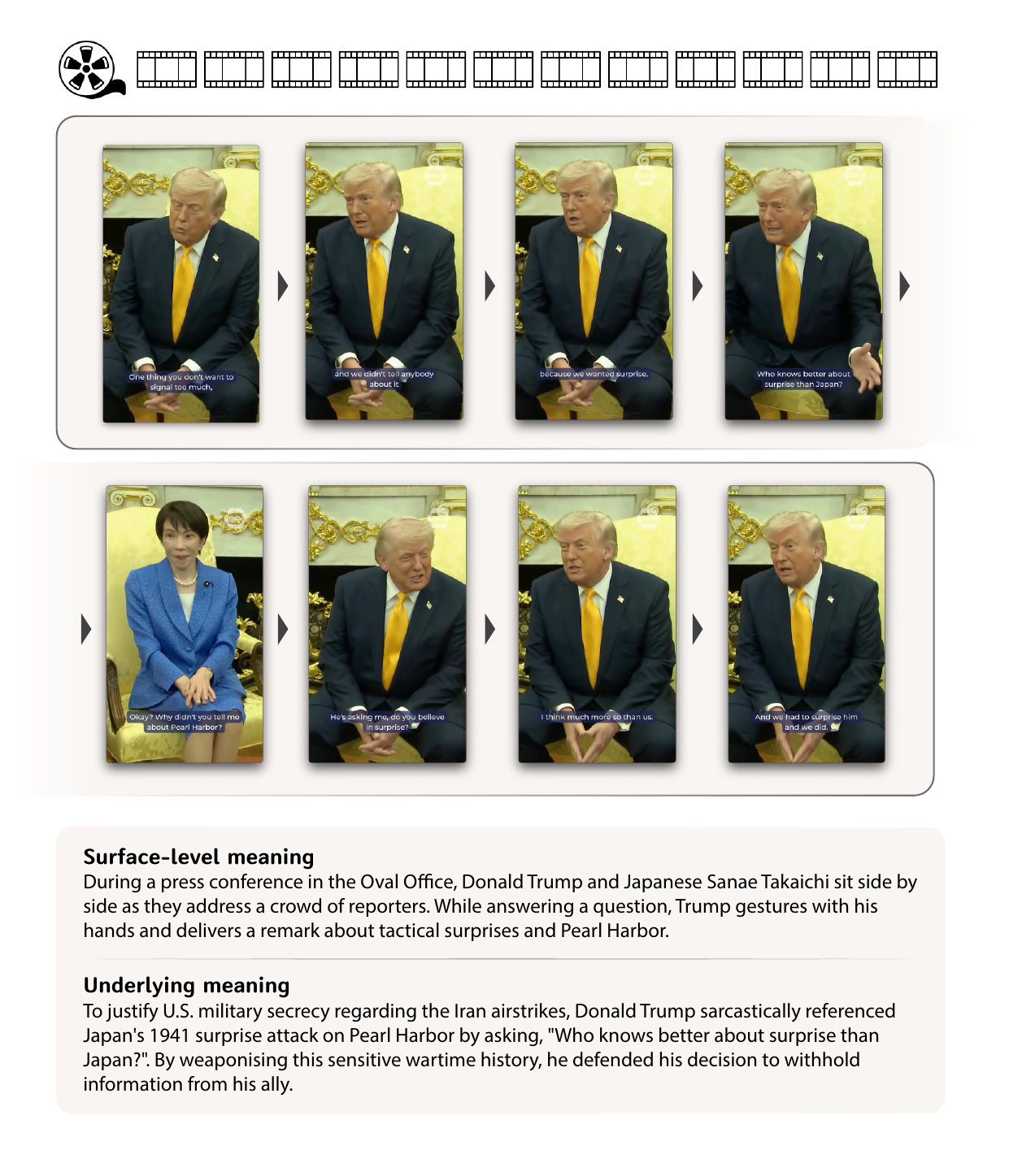}
    \caption{Example of a drivelological video requiring speech understanding, historical knowledge, and pragmatic inference.}
    \label{fig:example_audio}
\end{figure*}

\begin{figure*}[tp]
    \centering
    \includegraphics[width=0.95\linewidth]{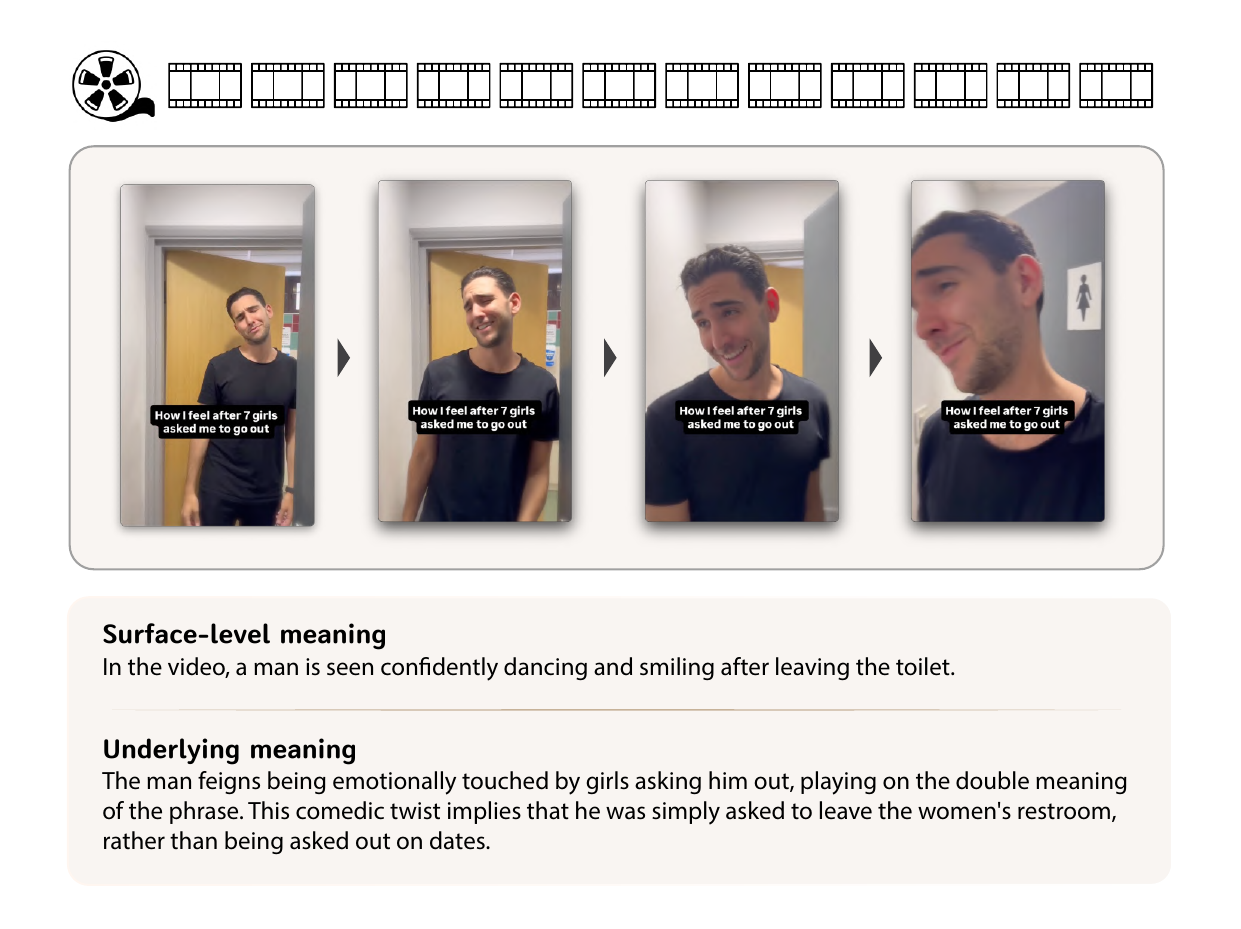}
    \caption{Example of a drivelological video requiring the interaction of on-screen text and visual context.}
    \label{fig:example_text_vision}
\end{figure*}

\begin{figure*}[tp]
    \centering
    \includegraphics[width=0.9\linewidth]{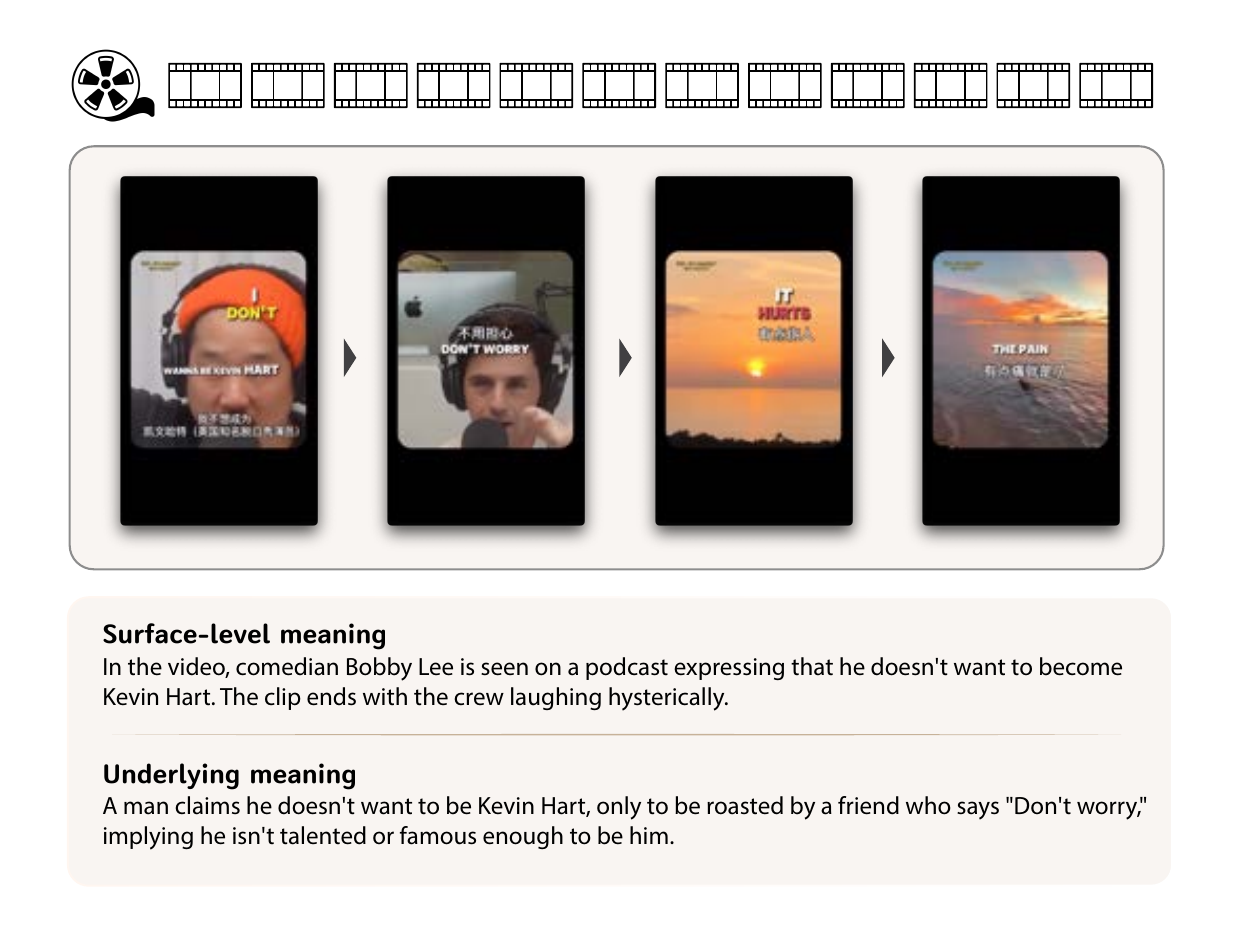}
    \caption{Example of a drivelological video requiring audio-visual pragmatic reasoning.}
    \label{fig:example_audio_vision}
\end{figure*}

\begin{figure*}[tp]
    \centering
    \includegraphics[width=0.9\linewidth]{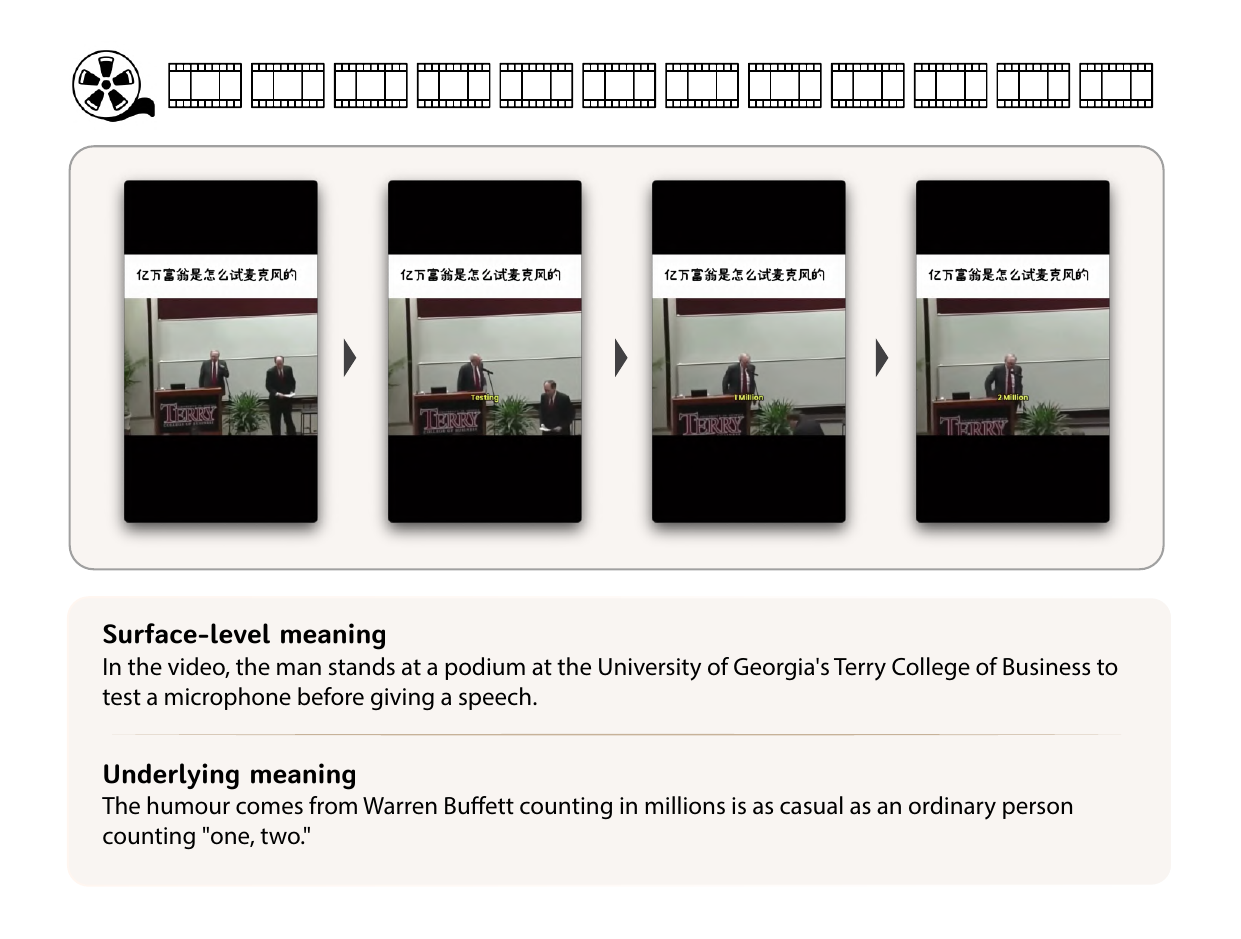}
    \caption{Example of a drivelological video requiring on-screen text, audio, and visual context.}
    \label{fig:example_text_audio_vision}
\end{figure*}

\end{document}